\documentclass{article}

\PassOptionsToPackage{numbers, compress}{natbib}

\usepackage[main, final]{neurips_2026}

\usepackage[utf8]{inputenc} 
\usepackage[T1]{fontenc}    
\usepackage{hyperref}       
\usepackage{url}            
\usepackage{booktabs}       
\usepackage{amsfonts}       
\usepackage{nicefrac}       
\usepackage{microtype}      
\usepackage{xcolor}         
\usepackage{amsmath}
\usepackage{makecell}
\usepackage{multirow} 
\usepackage{dsfont}
\usepackage{soul}
\usepackage{pifont}
\usepackage{mathtools}
\usepackage{graphicx}

\title{CRISP: Fixing Flying Pixels in Latent LiDAR Generation via Diffusion Decoding}

\author{
  Andrea Ceron$^{1,2}$ \And
  Michael Schmidt$^{2}$ \And
  Alvaro Marcos-Ramiro$^{2}$
  \AND
  Sebastian Schmidt$^{1,2}$ \And
  Benjamin Busam$^{1,3}$
  \AND
  $^{1}$TU München \quad $^{2}$BMW AG \quad $^{3}$Munich Center for Machine Learning \\
  \texttt{\{andrea.ceron, sebastian95.schmidt, b.busam\}@tum.de} \\
  \texttt{\{Michael.SE.Schmidt, Alvaro.Marcos-Ramiro\}@bmw.de}
}

\begin{document}

\maketitle

\begin{figure}[ht]
    \centering
    \includegraphics[width=0.99\linewidth]{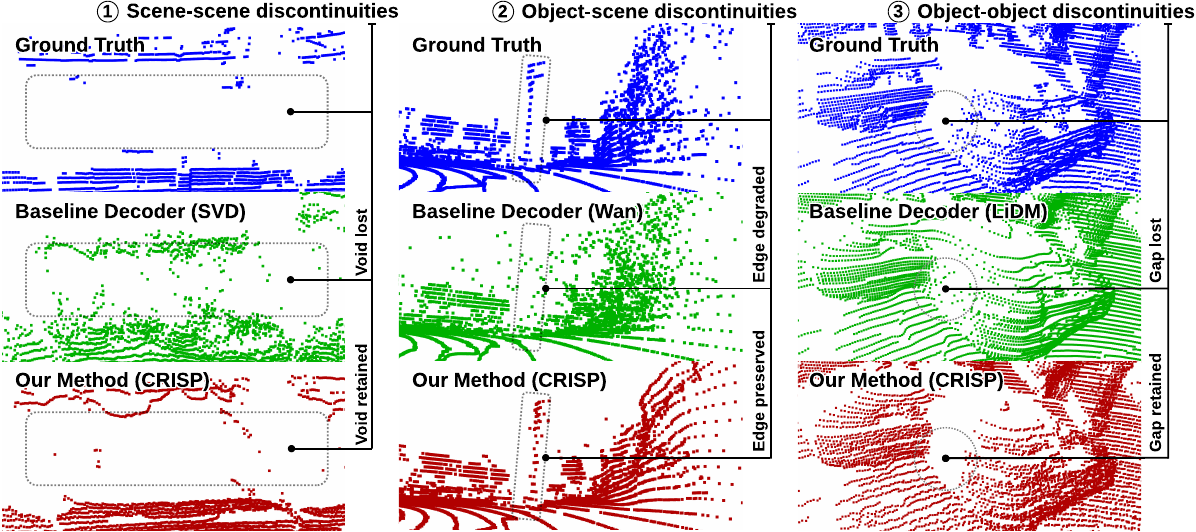}
    \caption{Qualitative comparison across three frozen latent backbones. \textcolor{blue}{Blue} denotes the ground-truth point cloud, \textcolor{green!50!black}{green} the inherited VAE decoder, and \textcolor{red!70!black}{red} our method (CRISP). The three columns show one failure per backbone: SVD, Wan2.1, and LiDM. Across all cases, CRISP suppresses flying pixels and boundary-bridging artifacts that persist in the baseline reconstructions, yielding cleaner geometry and sharper depth contours.}
    \label{fig:intro_figure}
\end{figure}

\begin{abstract}
Latent LiDAR pipelines suffer from flying pixels: convolutional VAEs blur sharp radial depth discontinuities, yielding edge depths that back-project to points floating between surfaces. We identify this as a major, directly correctable decoder bottleneck and introduce \textbf{CRISP}: a pixel-space diffusion decoder with a backbone-agnostic latent adapter, DiT-based denoiser, and support mask predictor. CRISP replaces video-VAE and LiDAR-native decoders alike while keeping the encoder and latent generator fixed. Across KITTI-360, SemanticKITTI, and nuScenes, replacing only the decoder reduces FSVD/FPVD by 50.4\% on average across frozen backbones; for generic video VAEs, the reductions reach 71\%/74\%. On the LiDAR-native LiDM backbone, FRID drops by 71\%, with the largest gains at depth discontinuities. In a pretrained LiDM world model, the same zero-shot replacement improves FSVD by 15.5\%, narrowing the sim-to-real gap.  Code and checkpoints will be released on the \href{https://andrea25512.github.io/CRISP/}{\textcolor{blue}{Project Page}}.

\end{abstract}

\section{Introduction}
LiDAR provides a geometric backbone for autonomous driving, underpinning scene perception, simulation, and safety-relevant planning~\cite{wang2025forging, yang2025methods, zhou2022review}. As these systems scale, synthetic LiDAR has become essential for coverage expansion and rare-scenario generation~\cite{kim2025synad}, and improvements in reconstruction fidelity translate directly to more trustworthy simulation for the safety-relevant perception and planning pipelines that depend on it. Modern pipelines address this by compressing LiDAR into a compact latent and generating within that space~\cite{rombach2022high, ran2024towards, hu2024rangeldm, zhao2026unidrivedreamer}, many of which decode through convolutional VAE decoders. These decoders fall into two families: general-purpose video VAEs~\cite{blattmann2023stable, wan2025wan} repurposed for LiDAR~\cite{zhao2026unidrivedreamer}, and LiDAR-native autoencoders~\cite{ran2024towards} that, like range-image diffusion models~\cite{zyrianov2022learning, nakashima2024lidar}, use circular convolutions to respect the sensor's $360^\circ$ azimuthal sweep.

The consequence is a concrete and recurring artifact: \emph{flying pixels}~\cite{xu2025pixel, liu2026l3dr}. Convolutional VAE decoders smooth across the sharp depth contours that define LiDAR geometry, emitting interpolated depth at object edges: a leakage problem so recognized that LiDM resorts to flat convolutional kernels specifically to limit it~\cite{ran2024towards}. Even this LiDAR-native design, with $1{\times}4$ kernels and horizontal-only resampling in its outermost curve-wise blocks and circular padding that closes the azimuthal seam, still convolves along each scan row and thus across radial depth discontinuities. When the blurred range map is back-projected to 3D, those interpolated boundary pixels become floating points anchored to neither surface~\cite{xu2025pixel}, clustering at object boundaries and thin structures, and propagating into downstream perception~\cite{liu2026l3dr}.

Pixel-Perfect Depth~\cite{xu2025pixel} provides a useful decoder-level diagnosis for geometric prediction: latent decoding can corrupt sharp structure even when the latent representation is informative. We ask whether this bottleneck becomes even more consequential for LiDAR, where range maps are sparse, anisotropic, and subject to explicit missing returns, making decoder oversmoothing especially damaging after back-projection~\cite{marcus2025physically}. The LiDAR generation literature has meanwhile advanced through better latents, conditioning, and pretrained or external priors~\cite{ran2024towards, xiang2025sg, sereyjol2026leveraging, zhao2026unidrivedreamer}, leaving one question that this paper answers directly: \emph{is the decoder itself a bottleneck, and can replacing it alone improve latent LiDAR reconstruction?}

We introduce \textbf{CRISP}, a plug-in diffusion decoder that directly addresses this question by replacing the decoder alone, producing crisp, geometrically faithful LiDAR reconstructions from any frozen latent backbone while preserving sharp depth boundaries and suppressing the flying-pixel artifacts that convolutional decoders introduce at depth discontinuities. 

Our contributions are:
\begin{itemize}
    \item We identify convolutional VAE decoders as a major geometry bottleneck causing oversmooth LiDAR maps with flying pixels at depth discontinuities, and show that this bottleneck is correctable: replacing the decoder alone recovers the boundary geometry that the frozen latent already carries.
    \item \textbf{CRISP} adapts pixel-space diffusion decoding~\cite{xu2025pixel} to the sparse, support-incomplete range-map regime LiDAR requires. It includes a backbone-agnostic latent adapter, a DiT denoiser with early fusion for geometry-first denoising, and a support mask predictor for sparse valid-return modeling.
    \item The gains are significant. Replacing only the decoder reduces FSVD/FPVD by 71\%/74\% for generic video VAEs and FRID by 71\% for the LiDAR-native backbone under frozen encoders, with edge F-score improving 107\% on average; inside a frozen LiDM~\cite{ran2024towards} world model, the same zero-shot swap improves FSVD by 15.5\% and FRID by 4.0\%, showing the bottleneck is actionable without retraining.
\end{itemize}

\section{Related Work}
\label{sec:related}
LiDAR synthesis has progressed from sensor-level simulation to learned range-view and latent generative models. Early work improved realism with real-world assets, ray casting, and learned sensor deviations~\cite{manivasagam2020lidarsim}, while deep generative models synthesized scans as structured 2D point maps~\cite{caccia2019deep}. Subsequent methods exploited spinning-LiDAR image geometry more directly: LiDARGen uses stochastic denoising for equirectangular LiDAR~\cite{zyrianov2022learning}, UltraLiDAR learns compact discrete representations for completion and generation~\cite{xiong2023learning}, and R2DM adapts DDPMs to range/reflectance images~\cite{nakashima2024lidar}. RangeLDM then moves this line into latent diffusion through range-view VAEs for efficient, realistic generation~\cite{hu2024rangeldm}. This motivates our setting: modern LiDAR generators increasingly depend on compressed range-map latents, yet leave the decoder largely unexamined and fixed.

Most recent LiDAR generation work improves latent modeling, conditioning, or the use of pretrained priors rather than the decoder itself. LiDM~\cite{ran2024towards} established a latent diffusion pipeline built on a convolutional VAE for LiDAR range maps. SG-LDM~\cite{xiang2025sg} adds latent semantic alignment for semantic-to-LiDAR synthesis while keeping a standard 2D diffusion/autoencoding setup. R3DPA~\cite{sereyjol2026leveraging} improves range-image generation through RGB-pretrained priors and 3D representation alignment while retaining the range-VAE paradigm. UniDriveDreamer~\cite{zhao2026unidrivedreamer} adapts a general-purpose pretrained VAE to LiDAR data through LiDAR-specific fine-tuning and cross-modal latent anchoring for joint RGB-LiDAR generation. Across these lines, the main advances come from \emph{better latents, conditioning, and priors,} not from replacing the convolutional decoder. Circular padding, as in LiDM and R2DM, resolves the azimuthal seam but not the \emph{radial} discontinuities that cause flying pixels: replacing only the LiDM decoder still lowers edge CD by 48.8\% and FRID by 71.4\% (Tables~\ref{tab:results_geometry},~\ref{tab:results_perceptual}).

A complementary line tackles geometry after or beyond decoding. L3DR~\cite{liu2026l3dr} attributes artifacts such as depth bleeding and wavy surfaces to range-view generation and corrects them with a post-hoc 3D residual rectifier, leaving the latent diffusion/decoding stack in place, while LidarDM~\cite{zyrianov2025lidardm} shifts the problem to 4D world generation and physics-based raycasting, prioritizing controllability and temporal coherence rather than decoder redesign. Both are complementary, but they \emph{do not isolate the decoder as a bottleneck}; our focus is precisely that stage, replacing the convolutional decoder itself, inherited or LiDAR-native, with a plug-in pixel-space diffusion decoder tailored to sparse LiDAR range maps.

LiDAR makes the decoder bottleneck strictly harder than in monocular depth: range maps are sparse, anisotropic, and subject to explicit missing returns~\cite{marcus2025physically}, so decoder-induced smoothing is more damaging, not less. Pixel-Perfect Depth~\cite{xu2025pixel} shows that VAE compression degrades edge sharpness even when reconstructing \emph{ground-truth} depth, confirming latent decoding as a geometric bottleneck independent of the generative model. It is our closest architectural relative and also denoises in pixel space with cascaded diffusion transformers~\cite{peebles2023scalable}, but it conditions on a persistent clean RGB image rather than a frozen latent, predicts dense depth without a support branch, and outputs a single depth map rather than depth composed with a validity mask; ablating the first two differences degrades CRISP measurably (Table~\ref{tab:ablation_dit}).

PiD~\cite{lu2026pid} and $\epsilon$-VAE~\cite{zhao2025epsilon} replace convolutional decoders with diffusion decoders for natural images, where adding structure or reconstructing stochastically is acceptable; LiDAR instead requires per-frame faithfulness, so CRISP regresses depth against ground truth on valid pixels rather than sampling freely. RAE~\cite{zheng2025diffusion} places the bottleneck at the other end: it replaces the VAE encoder with a frozen representation encoder, which requires a new decoder and a retrained generator. CRISP places it at the decoder, so it keeps both encoder and latent generator frozen, which enables a zero-shot decoder swap within a frozen LiDM world model (Table~\ref{tab:results_world_model}).

CRISP is designed around the decoder failure mode as it manifests in LiDAR, where the consequences after back-projection are more severe; full numbers for all three contrasts, additional coverage of architecture-level structural responses, downstream applications of decoder quality, and the pixel-space diffusion lineage are in Appendix~\ref{sec:related_appendix}.

\section{Method}
\begin{figure}[t]
    \centering
    \includegraphics[width=\linewidth]{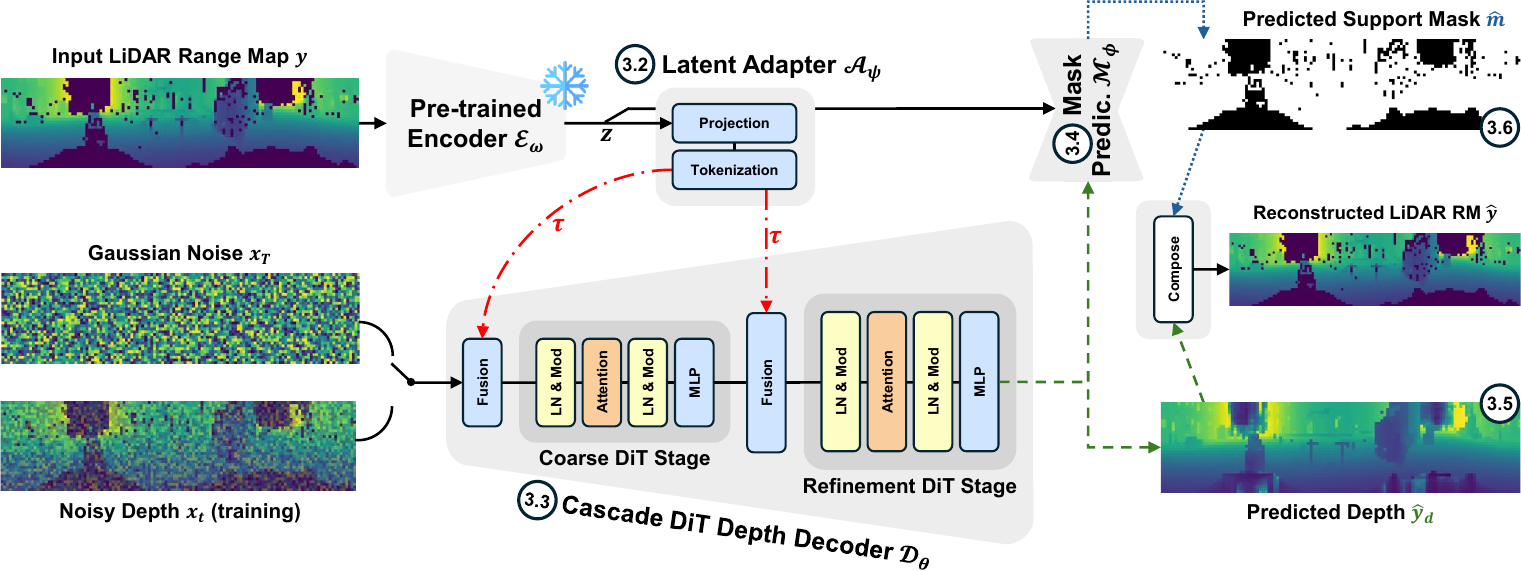}
    \caption{Architecture of CRISP. \textit{Left:} the frozen encoder $\mathcal{E}_\omega$ maps the input range map $y$ to latent $z$. \textit{Top:} the latent adapter $\mathcal{A}_{\psi}$ projects and tokenizes $z$ into conditioning tokens \textcolor{red!70!black}{$\tau$}; subsequently, the support mask predictor $\mathcal{M}_{\phi}$ takes $z$ and \textcolor{green!50!black}{$\hat{y}_d$} to produce the binary mask \textcolor{blue}{$\hat{m}$}. \textit{Centre:} the cascade DiT decoder $\mathcal{D}_{\theta}$ denoises $x_t$ on $\tau$ at two fusion points (red dashed arrows), yielding the dense depth \textcolor{green!50!black}{$\hat{y}_d$}. \textit{Right:} $\hat{m}$ and $\hat{y}_d$ are composed as $\hat{y} = \hat{m} \odot \hat{y}_d - (1-\hat{m})$ to produce the final LiDAR range map.}
    \label{fig:architecture}
\end{figure}

Convolutional VAE decoders, whether repurposed video decoders or LiDAR-native, smooth over the sharp support discontinuities that define LiDAR geometry, generating flying pixels at object boundaries that persist independent of latent quality. To address this directly, CRISP introduces a two-branch pixel-space reconstruction module tailored to LiDAR geometry, even when keeping the pretrained encoder frozen. Given an input LiDAR range map, the encoder produces a latent representation shared by both branches (Fig.~\ref{fig:architecture}). A \emph{latent adapter} (Sec.~\ref{sec:method:adapter}) converts the encoder output into conditioning tokens for the denoising transformer. The first reconstruction branch is a \emph{DiT-powered pixel diffusion decoder} (Secs.~\ref{sec:method:dit},~\ref{sec:method:diffloss}) that reconstructs a dense depth map $\hat{y}_d$ by denoising a Gaussian noise sample conditioned on the adapted latent tokens~\cite{salimans2022progressive}. Conditioned on both $z$ and $\hat{y}_d$, a \emph{support mask predictor} (Secs.~\ref{sec:method:mask},~\ref{sec:method:maskloss}) then identifies which pixels contain valid LiDAR returns. The final sparse LiDAR output is $\hat{y} = \hat{m} \odot \hat{y}_d - (1-\hat{m})$.

\subsection{Encoding Setup}
\label{sec:method:notation}

Let $y \in \mathbb{R}^{H \times W}$ be a LiDAR depth image in cylindrical range-map form~\cite{wu2018squeezeseg} and $m \in \{0,1\}^{H \times W}$ the corresponding binary mask, where $m_{ij}=1$ indicates a valid LiDAR return at pixel $(i,j)$.  Following prior range-map LiDAR work~\cite{ran2024towards, nakashima2024lidar, xu2025u4d, zyrianov2022learning, kirby2024logen}, we operate in a log-normalised depth representation (Appendix~\ref{app:data-preprocessing}, Eqs.~\ref{eq:kitti-norm}-\ref{eq:nus-norm}), mapping $\tilde{y}\in[-1,1]^{H\times W}$ to allocate more resolution to near-range structure~\cite{wu2018squeezeseg}. A pretrained encoder $\mathcal{E}_\omega$ maps $\tilde{y}$ to a latent tensor $z = \mathcal{E}_\omega(\tilde{y})$~\cite{kingma2013auto, hu2024rangeldm}.

\subsection{Latent Adapter}
\label{sec:method:adapter}

CRISP supports heterogeneous latent backbones, including LiDAR-native and general-purpose video VAEs (e.g.\ LiDM~\cite{ran2024towards}, SVD~\cite{blattmann2023stable},
or Wan2.1~\cite{wan2025wan}). Since these encoders produce latents with different channel counts and spatial resolutions, the adapter $\mathcal{A}_{\psi}$ maps them to a common token sequence:
\begin{equation}
  \tau = \mathcal{A}_{\psi}(z)
  =
  \mathbf{W}_{s}\,
  \mathrm{Flat}\!\left(\mathbf{W}_{c} \cdot z\right),
  \label{eq:adapter}
\end{equation}
where $\mathbf{W}_{c}$ is a pointwise channel projection (applied independently at each spatial location) and $\mathbf{W}_{s}$ reduces the flattened spatial token count to the DiT conditioning length. This separates the decoder from the latent geometry of any specific backbone.

\subsection{DiT-Powered Pixel Diffusion Decoder}
\label{sec:method:dit}

The denoiser $\mathcal{D}_{\theta}$ is a Diffusion Transformer (DiT) operating directly in LiDAR pixel space~\cite{peebles2023scalable,xu2025pixel}. We use a standard Gaussian forward process and velocity parameterization~\cite{ho2020denoising,song2020score,salimans2022progressive}, where $\alpha_t$ and $\sigma_t$ are the signal and noise coefficients of the diffusion schedule ($\alpha_t^2+\sigma_t^2=1$):
\begin{equation}
  x_t = \alpha_t \tilde{y} + \sigma_t \epsilon,
  \qquad
  v^\star \coloneqq \alpha_t \epsilon - \sigma_t \tilde{y},
  \qquad
  \hat{v} \coloneqq \mathcal{D}_{\theta}(x_t,\tau,t),
  \label{eq:diffusion}
\end{equation}
with $\epsilon\sim\mathcal{N}(0,I)$, $v^\star$ the ground-truth velocity target and $\hat{v}$ the predicted velocity. The dense depth estimate $\hat{y}_d$ is recovered as
\begin{equation}
  \hat{y}_d = \alpha_t x_t - \sigma_t \hat{v}.
  \label{eq:x0_from_v}
\end{equation}

\paragraph{Early and mid-level latent fusion.} The key architectural change is how latent information enters the denoiser. Unlike monocular depth diffusion, where a guidance image provides clean structure throughout the network~\cite{xu2025pixel}, our decoder starts from pure Gaussian noise. We therefore inject latent tokens both immediately after patch embedding and again at the network midpoint:
\begin{equation}
  h_0 = \mathrm{Fuse}_{\mathrm{early}}(\mathrm{Tok}(x_t),\tau),
  \qquad
  h_{K/2} \leftarrow \mathrm{Fuse}_{\mathrm{mid}}
  \left(\mathrm{Block}_{1:K/2}(h_0,t),\tau\right),
  \label{eq:fusion}
\end{equation}
after which the remaining DiT blocks predict $\hat{v}$ (Eq.~\ref{eq:diffusion}). Each fusion module concatenates range-map tokens and latent tokens and projects them back to the hidden dimension. This early conditioning is critical because the denoiser must recover global scene layout before refining local geometry~\cite{xu2025pixel}. Full architectural specifications are given in Appendix~\ref{sec:arch:dit}.

\subsection{Support Mask Predictor}
\label{sec:method:mask}

LiDAR range maps contain two coupled signals: the returned range values and the binary support indicating where the sensor received a valid return. We follow the standard ray-drop formulation (Appendix~\ref{sec:related_appendix}), but use the support predictor as a \emph{latent-conditioned branch of the plug-in decoder}. Unlike standalone ray-drop heads, $\mathcal{M}_{\phi}$ operates directly on the frozen encoder latent $z$ and predicts the valid-return support used to sparsify the decoded range map.

The branch upsamples $z$ to the range-map resolution and outputs a logit field $r_{\phi}$, whose sigmoid defines a Bernoulli probability map:
\begin{equation}
  r_{\phi} = \mathcal{M}_{\phi}(z, \hat{y}_d),
  \qquad
  \pi_{\phi} = \sigma(r_{\phi}),
  \qquad
  q_{\phi}(\hat{m}\mid z, \hat{y}_d)
  =
  \prod_{i,j}
  \mathrm{Bernoulli}\!\left(\hat{m}_{ij};\,\pi_{\phi,ij}\right).
  \label{eq:mask_logits}
\end{equation}
During training, support losses are applied to $r_{\phi}$ and $\pi_{\phi}$ against the ground-truth mask $m$. At inference, the Bernoulli field yields a binary support mask that sparsifies the dense depth:
\begin{equation}
  \hat{m} \sim q_{\phi}(\hat{m}\mid z, \hat{y}_d),
  \qquad
  \hat{y} = \hat{m} \odot \hat{y}_d - (1-\hat{m}).
  \label{eq:mask_inference}
\end{equation}
Confident regions ($\pi_{\phi,ij}\!\approx\!1$ or $\approx\!0$) are effectively deterministic, while ambiguous structures such as vegetation and thin objects can produce stochastic ray-drop variation across samples. The U-Net architecture and dual-branch conditioning are detailed in Appendix~\ref{sec:arch:mask}.

\subsection{Geometry-Aware Diffusion Objective}
\label{sec:method:diffloss}

Both branches share one structure: a \emph{region objective} on valid pixels, \emph{spatial regularisers} of first and second differential order, and \emph{optimisation machinery} that shapes training without modelling a separate failure mode (per-term failure mode and provenance in Appendix~\ref{sec:loss_justification}).

The depth branch minimizes the following composite objective:
\begin{equation}
  \mathcal{L}_{\mathrm{depth}}
  =
  \mathcal{L}_{\mathrm{vel}} 
  + \mathcal{L}_{\mathrm{der}}
  + \lambda_{\mathrm{grad}}\mathcal{L}_{\mathrm{grad}} 
  + \mathcal{L}_{\mathrm{vel,raw}} 
  + \lambda_{x0}\mathcal{L}_{x0}
  \qquad \lambda_{\mathrm{grad}},\lambda_{x0} \geq 0.
  \label{eq:depth_loss}
\end{equation}
The region objective $\mathcal{L}_{\mathrm{vel}}$ is a boundary-aware velocity regression on valid pixels (importance-weighted). The spatial regularisers are $\mathcal{L}_{\mathrm{der}}$, which penalises finite-difference residuals along both spatial axes, and $\mathcal{L}_{\mathrm{grad}}$, a multi-scale first-order gradient loss on $\hat{y}_d$ from LiDM~\cite{ran2024towards}. The optimisation machinery is $\mathcal{L}_{\mathrm{vel,raw}}$, the unweighted counterpart of $\mathcal{L}_{\mathrm{vel}}$ providing a stable baseline signal, and $\mathcal{L}_{x0}$, a direct $\ell_1$ penalty on $\hat{y}_d$ concentrated at low noise levels. Definitions are in Appendix~\ref{sec:arch:diffloss}; weights are in Table~\ref{tab:loss-weights}. 

\paragraph{Valid-only structure-aware importance weighting.} \label{par:valid_only_structure_weighting} LiDAR errors are not equally costly: mistakes at depth discontinuities back-project to large 3D deviations, while errors in flat regions are comparatively harmless. A naïve Sobel or Laplacian on a zero-filled sparse map produces spurious responses at valid/void borders; we avoid this via validity-normalised filtering~\cite{knutsson1993normalized, eldesokey2020uncertainty, uhrig2017sparsity}:
\begin{equation}
  \bar{y}
  =
  \frac{G_{\sigma} * (m \odot \tilde{y})}
       {G_{\sigma} * m + \epsilon},
  \label{eq:valid_blur}
\end{equation}
where $*$ denotes convolution, $\odot$ element-wise multiplication, and $G_\sigma$ a
Gaussian kernel ($5{\times}5$ kernel, $\sigma{=}2.0$). We compute $h=|\tilde{y}-\bar{y}|$ and map it to a bounded importance
weight:
\begin{equation}
  w = m \odot \Bigl(1 + (\lambda_{\max}-1)\,\mathrm{Norm}(h)\Bigr),
  \quad \lambda_{\max} = 1.5.
  \label{eq:importance}
\end{equation}
Flat valid regions receive weight $1$; pixels near true depth discontinuities receive weight
up to $\lambda_{\max}$. The boundary-aware loss $\mathcal{L}_{\mathrm{vel}}$ reuses
Eq.~\ref{eq:vel_raw} with $m$ replaced by $w$ (Appendix~\ref{sec:arch:diffloss}).

\subsection{Support Mask Objective}
\label{sec:method:maskloss}

The mask branch minimizes:
\begin{equation}
  \mathcal{L}_{\mathrm{mask}}
  =
  \lambda_{\mathrm{focal}}\mathcal{L}_{\mathrm{focal}}
  + \lambda_{\mathrm{dice}}\mathcal{L}_{\mathrm{dice}}
  + \lambda_{\mathrm{edge}}\mathcal{L}_{\mathrm{edge}}
  + \lambda_{\mathrm{lap}}\mathcal{L}_{\mathrm{lap}}
  + \lambda_{\mathrm{hard}}\mathcal{L}_{\mathrm{hard}}
  + \lambda_{\mathrm{ds}}\mathcal{L}_{\mathrm{ds}},
  \qquad\lambda_\ast \geq 0.
  \label{eq:mask_loss}
\end{equation}
     
Valid LiDAR support is a multi-scale prediction problem: the mask must capture coarse global structure, sharp foreground-background boundaries, thin valid returns, and isolated single-pixel ray-drop holes. No single loss handles all four failure modes. As the region objective, focal BCE~\cite{lin2017focal} addresses class imbalance and focuses the gradient on hard boundary and ray-drop regions, and soft Dice~\cite{milletari2016v} optimizes overlap directly, independent of pixel counts. As spatial regularisers, a first-order edge term enforces boundary alignment~\cite{huang2019indoor}, while a second-order Laplacian term~\cite{ma2019self,zhong2019deep} captures isolated ray-drop events invisible to first-order penalties. As optimisation machinery, contextual hard-pixel mining~\cite{shrivastava2016training} up-weights spatially fragile regions by target-neighborhood structure, and deep supervision~\cite{lee2015deeply,qin2019basnet} at two intermediate resolutions ensures the gradient signal reaches shallow layers. 
Per-term failure modes and provenance are in Table~\ref{tab:loss-provenance}, weights in Table~\ref{tab:mask-weights}. Per-branch ablations identify a seven-term reduced objective that drops the overlapping regulariser of each branch and the mask machinery: removing $\mathcal{L}_{\mathrm{grad}}$ raises depth val L1 by only 2.9\% on nuScenes and 0.1\% on KITTI-family (Table~\ref{tab:depth-loo}), and Focal~+~Dice~+~Laplacian retains 96.7\% of the six-term IoU gain over focal-only on LiDM/KITTI-family (Table~\ref{tab:mask-minimal}).

\subsection{Full Objective and Encoder Adaptation}
\label{sec:method:full}

The default CRISP freezes the encoder and trains only the plug-in decoder branches, and optimizes:
\begin{equation}
  \mathcal{L}
  =
  \mathcal{L}_{\mathrm{depth}}
  + \lambda_{\mathrm{mask}}\mathcal{L}_{\mathrm{mask}},
  \qquad \lambda_{\mathrm{mask}} \geq 0.
  \label{eq:full_loss}
\end{equation}
Weights are set once per training stage by equalising initial term magnitudes, a static simplification of GradNorm~\cite{chen2018gradnorm}, and reused unchanged for every backbone and beam count. We further evaluate encoder-adapted SVD and Wan2.1 variants~\cite{blattmann2023stable,wan2025wan}, excluding LiDM since its encoder is LiDAR-native~\cite{ran2024towards}. In the baseline setting, the original VAE architecture is adapted to single-channel LiDAR using the UniDriveDreamer~\cite{zhao2026unidrivedreamer} objective, without modification (Appendix~\ref{app:encoder-adapt}). In the CRISP setting, we instead unlock the encoder and jointly train it with the CRISP decoder, adding a KL regulariser to keep the encoder posterior close to the prior $p(z)=\mathcal{N}(0,I)$:
\begin{equation}
  \mathcal{L}
  =
  \mathcal{L}_{\mathrm{depth}}
  + \lambda_{\mathrm{mask}}\mathcal{L}_{\mathrm{mask}}
  + \lambda_{\mathrm{KL}}\mathrm{KL}\!\left(q_\omega(z \mid \tilde{y})\,\|\,p(z)\right),
  \qquad \lambda_{\mathrm{mask}},\lambda_{\mathrm{KL}} \geq 0
  \label{eq:full_loss_adapt}
\end{equation}

\section{Experimental Setup}
\paragraph{Datasets and evaluation regimes.}
We evaluate on three standard automotive LiDAR benchmarks: KITTI-360~\cite{liao2022kitti}, SemanticKITTI~\cite{behley2019semantickitti}, and nuScenes~\cite{caesar2020nuscenes}. Following LiDM~\cite{ran2024towards}, KITTI-360 and SemanticKITTI are grouped as the KITTI family; nuScenes provides a larger-scale benchmark with lower vertical LiDAR resolution. 

\paragraph{Latent backbones and comparison settings.}
We evaluate CRISP with three latent VAE backbones: LiDM, a LiDAR-native autoencoder with circular padding, evaluated on the KITTI-family setting~\cite{ran2024towards}; and SVD and Wan2.1, two general-purpose video VAEs evaluated on nuScenes~\cite{blattmann2023stable,wan2025wan}. We include Wan2.1 as a realistic stress test for inherited video-VAE decoders used in recent driving world models~\cite{zhao2026unidrivedreamer}, not as an artificially weak baseline. Because SVD and Wan expect RGB input, we study two regimes for them: \emph{frozen replacement} (\textbf{$\times$3}), where the encoder is fixed and the depth channel is replicated three times to match the pretrained (Appendix~\ref{app:frozen-backbones}); and \emph{encoder adaptation} (\textbf{$\times$1}), where encoders are either independently or jointly finetuned with CRISP for LiDAR, and the first/last VAE layers are made single-channel (Appendix~\ref{app:encoder-adapt}). For LiDM, whose encoder natively accepts single-channel depth, we additionally evaluate \emph{plug-and-play world-model decoding}, replacing the decoder inside a pretrained LiDM world model while keeping generated latents and latent dynamics unchanged.

\paragraph{Metrics.}
We evaluate perception realism, distributional fidelity, and paired 3D reconstruction. For perception quality, we report LiDM's Fréchet-style metrics: FRID, FSVD, and FPVD~\cite{ran2024towards}. Analogous to FID~\cite{heusel2017gans}, these compare real and reconstructed LiDAR in feature spaces: range-image features for FRID, sparse-volume features for FSVD, and point-volume features for FPVD. 
For distributional fidelity, we report Jensen-Shannon Divergence (JSD), Earth Mover's Distance (EMD), and Minimum Matching Distance (MMD), point-cloud generation metrics that compare occupancy statistics and nearest-set matching~\cite{achlioptas2018learning}. For paired geometry, we report Chamfer Distance (CD) and F-score@0.5m after back-projection to 3D~\cite{fan2017point,tatarchenko2019single}. CD and EMD measure point-set alignment, while F-score summarizes precision and recall under a fixed distance threshold.

\paragraph{Boundary-focused geometric analysis.}
Since CRISP targets decoder artifacts at depth discontinuities, we report CD and F-score not only on the full reconstructed point cloud but also on two ground-truth-defined subsets. Following the valid-only structure signal used for boundary-aware supervision in Sec.~\ref{par:valid_only_structure_weighting}, we identify edge pixels by thresholding local depth discontinuities computed over valid LiDAR returns (by value $0.1$). The \emph{edge} subset contains valid pixels above this threshold, while the \emph{smooth} subset contains the remaining valid pixels. Invalid/empty transitions are ignored, so missing returns are not counted as edges. Each subset is back-projected separately before evaluation. Full training and evaluation setup is provided in Appendix~\ref{app:training-protocol}. Asset licenses are listed in Appendix~\ref{app:licenses}.

\section{Results}
Because no single metric captures boundary geometry, we emphasize trends consistent across perceptual, paired-3D, and edge/smooth metrics, and report distributional regressions explicitly.
\begin{table*}[t]
\centering
\small
\setlength{\tabcolsep}{3.8pt}
\caption{Statistical and perceptual reconstruction results. \emph{base.}~uses the original decoder; \emph{ours}~replaces it with CRISP under the same encoder setting. Best values within each backbone pair are bold. Mean over five runs; parenthesized digits give std at mean precision (e.g.\ $0.123\,(45)$ $= 0.123 \pm 0.045$; $(00)$ denotes std ${<}0.001$ at that precision). 
FRID requires 64-beam range-image features and is reported only for the KITTI-family (LiDM) setting. $\downarrow$/$\uparrow$: lower/higher is better. \textbf{Same protocol for all tables.}}
\label{tab:results_perceptual}
\vspace{0.5em}
\begin{tabular}{llrrrrrr}
\toprule
\multirow{2}{*}{Setting} & \multirow{2}{*}{Dec.}
& \multicolumn{3}{c}{Statistical}
& \multicolumn{3}{c}{Perceptual} \\
\cmidrule(lr){3-5}\cmidrule(lr){6-8}
&& JSD$\downarrow$ & EMD$\downarrow$ & MMD$\downarrow{\times}10^{-4}$
& FSVD$\downarrow$ & FPVD$\downarrow$ & FRID$\downarrow$ \\
\midrule
\multicolumn{8}{l}{\textbf{Frozen backbones}} \\
SVD~$\times$3
& base. & 0.18961 (00) & 0.25550 (00) & 2.6946 (44) & 156.487 (23) & 153.599 (20) & -- \\
& ours     & \textbf{0.07145 (02)} & \textbf{0.07653 (01)} & \textbf{2.2505 (20)} & \textbf{40.244 (99)} & \textbf{36.847 (75)}  & -- \\
\addlinespace[2pt]

Wan~$\times$3
& base. & 0.16373 (00) & 0.11840 (00) & 2.2505 (00) & 100.917 (00) & 113.381 (00) & -- \\
& ours     & \textbf{0.07326 (02)} & \textbf{0.08871 (03)} & \textbf{1.7193 (41)} & \textbf{32.534 (40)} & \textbf{31.065 (29)} & -- \\
\addlinespace[2pt]

LiDM
& base. & 0.04975 (00) & 0.53762 (00) & 0.7721 (00) & 20.204 (00) & \textbf{16.154 (00)} & 2.2351 (00)\\
& ours     & \textbf{0.04902 (01)} & \textbf{0.51924 (24)} & \textbf{0.6964 (57)} & \textbf{16.693 (49)} & 17.034 (30) & \textbf{0.6389 (19)} \\
\midrule
\multicolumn{8}{l}{\textbf{Encoder-adapted backbones}} \\
SVD~$\times$1
& base. & 0.15660 (04) & 0.19354 (08) & 2.538\phantom{0} (10) & \textbf{40.813 (64)} & \textbf{41.775 (55)} & -- \\
& ours     & \textbf{0.08316 (06)} & \textbf{0.13930 (03)} & \textbf{2.0734 (92)} & 64.86\phantom{0} (19) & 52.83\phantom{0} (15) & -- \\
\addlinespace[2pt]

Wan~$\times$1
& base. & \textbf{0.06962 (00)} & \textbf{0.04700 (00)} & 1.3645 (00) & 24.029 (00) & 23.681 (00) & -- \\
& ours     & 0.08244 (00) & 0.14736 (00) & \textbf{1.1682 (00)} & \textbf{8.290 (21)} & \textbf{7.112 (18)} & -- \\
\bottomrule
\end{tabular}
\end{table*}

\begin{table*}[t]
\centering
\small
\setlength{\tabcolsep}{3.8pt}
\caption{Geometric breakdown into whole-cloud, edge, and smooth sub-clouds. CD$\downarrow$ and F-score@0.5m$\uparrow$ are reported for each region.} 
\label{tab:results_geometry}
\vspace{0.5em}
\begin{tabular}{llrrrrrr}
\toprule
\multirow{2}{*}{Setting} & \multirow{2}{*}{Dec.}
& \multicolumn{2}{c}{Whole}
& \multicolumn{2}{c}{Edge}
& \multicolumn{2}{c}{Smooth} \\
\cmidrule(lr){3-4}\cmidrule(lr){5-6}\cmidrule(lr){7-8}
&& CD$\downarrow$ & F$\uparrow$
& CD$\downarrow$ & F$\uparrow$
& CD$\downarrow$ & F$\uparrow$ \\
\midrule
\multicolumn{8}{l}{\textbf{Frozen backbones}} \\
SVD~$\times$3
& base. & 1.33107 (00) & 0.67804 (00) & 8.691 (00) & 0.0952 (00) & 1.18399 (03) & 0.73018 (00) \\
& ours     & \textbf{0.63967 (17)} & \textbf{0.80465 (02)} & \textbf{7.364 (00)} & \textbf{0.2383 (00)} & \textbf{0.53709 (09)} & \textbf{0.82158 (02)} \\
\addlinespace[2pt]

Wan~$\times$3
& base. & 0.77027 (00) & 0.73240 (00) & \textbf{5.645 (00)} & 0.1038 (00) & 0.59749 (00) & 0.79303 (00) \\
& ours     & \textbf{0.46453 (09)} & \textbf{0.82551 (01)} & 5.861 (05) & \textbf{0.2848 (02)} & \textbf{0.39544 (06)} & \textbf{0.84166 (01)} \\
\addlinespace[2pt]

LiDM
& base. & 0.16026 (00) & 0.93935 (00) & 7.913 (00) & 0.3781 (00) & 0.12358 (00) & 0.94477 (00) \\
& ours     & \textbf{0.14245 (27)} & \textbf{0.94323 (01)} & \textbf{4.052 (10)} & \textbf{0.4792 (02)} & \textbf{0.11032 (17)} & \textbf{0.94901 (01)} \\
\midrule
\multicolumn{8}{l}{\textbf{Encoder-adapted backbones}} \\
SVD~$\times$1
& base. & 0.81622 (40) & 0.72676 (07) & 8.248 (03) & 0.1151 (00) & 0.67092 (38) & 0.75422 (07) \\
& ours     & \textbf{0.66981 (30)} & \textbf{0.81795 (05)} & \textbf{6.009 (01)} & \textbf{0.2082 (00)} & \textbf{0.56524 (15)} & \textbf{0.83863 (05)} \\
\addlinespace[2pt]

Wan~$\times$1
& base. & 0.30994 (00) & 0.86691 (00) & 6.150 (00) & 0.2226 (00) & \textbf{0.24295 (00)} & 0.88618 (00) \\
& ours     & \textbf{0.29017 (01)} & \textbf{0.87341 (01)} & \textbf{2.485 (01)} & \textbf{0.4456 (01)} & 0.24362 (03) & \textbf{0.89043 (01)} \\
\bottomrule
\end{tabular}
\end{table*}

\begin{table*}[t]
\centering
\small
\setlength{\tabcolsep}{4pt}
\caption{Plug-and-play decoding inside a frozen LiDM world model: statistical, perceptual and geometric quality. The latent generator and sampled latents are fixed; only the decoder changes. Values with a decimal point inside std parentheses are written explicitly at the displayed precision.}
\label{tab:results_world_model}
\vspace{0.5em}
\begin{tabular}{lrrrrrr}
\toprule
\multirow{2}{*}{Decoder}
& \multicolumn{3}{c}{Statistical}
& \multicolumn{3}{c}{Perceptual} \\
\cmidrule(lr){2-4}\cmidrule(lr){5-7}
& JSD$\downarrow$ & EMD$\downarrow$ & MMD$\downarrow{\times}10^{-4}$
& FSVD$\downarrow$ & FPVD$\downarrow$ & FRID$\downarrow$ \\
\midrule
baseline & \textbf{0.21181 (87)} & 2.453 (13) & 3.980 (44) & 37.70\phantom{0} (52) & 28.85 (34) & 137.81 (4.79) \\
ours     & 0.21436 (83) & \textbf{2.420 (13)} & \textbf{3.849 (64)} & \textbf{31.871 (34)} & \textbf{28.11 (18)}  & \textbf{132.36 (4.59)}\\
\bottomrule
\end{tabular} \par
\vspace{0.3em}
\begin{tabular}{lrrrrrr}
\multirow{2}{*}{Decoder}
& \multicolumn{2}{c}{Whole Geometry}
& \multicolumn{2}{c}{Edge Geometry}
& \multicolumn{2}{c}{Smooth Geometry} \\
\cmidrule(lr){2-3}\cmidrule(lr){4-5}\cmidrule(lr){6-7}
& CD$\downarrow$ & F$\uparrow$
& CD$\downarrow$ & F$\uparrow$
& CD$\downarrow$ & F$\uparrow$ \\
\midrule
baseline & 16.21 (23) & 0.4484 (13) & 96.41 (1.33) & \textbf{0.0486 (17)} & 15.80 (22) & 0.4489 (13) \\
ours     & \textbf{15.85 (23)} & \textbf{0.4510 (13)} & \textbf{79.55 (1.09)} & 0.0389 (14) & \textbf{15.53 (23)} & \textbf{0.4514 (13)} \\
\bottomrule
\end{tabular}
\end{table*}

\paragraph{Frozen backbones: the decoder is the bottleneck.}
Replacing only the inherited decoder yields large improvements in the main perceptual and geometric metrics (Tables~\ref{tab:results_perceptual} and~\ref{tab:results_geometry}), with the only perceptual exception being a small FPVD regression on the LiDM backbone. Averaged over the three frozen settings, FSVD decreases by 53\% and FPVD by 48\%; for the two general-purpose video VAEs, where the modality mismatch is greatest, these averages reach 71\% and 74\%, reflecting how severely a decoder trained on dense natural images fails when confronted with sparse, anisotropic LiDAR geometry (Figs.~\ref{fig:qual_extra_svd3x},~\ref{fig:qual_extra_wan3x}). On the LiDAR-native LiDM backbone, FRID drops from 2.24 to 0.64 (71\%) showing that even the LiDAR-native decoder suppresses structure already present in the latent (Fig.~\ref{fig:qual_extra_lidm}). Whole-cloud CD decreases by 34\% on average (46\% for generic VAEs, 11\% for LiDM), the smaller LiDM gain consistent with a LiDAR-native encoder near its reconstruction ceiling, and the improvement scaling with modality gap confirms the decoder as a controllable bottleneck.

\paragraph{Geometric breakdown: boundary evidence.}
The sub-cloud breakdown is the sharpest test of the decoder-bottleneck hypothesis. Edge F-score improves by an average of 107\% across all five settings versus 9\% for the whole cloud; a roughly 12$\times$ contrast that localizes the failure mode: convolutional oversmoothing is not a diffuse reconstruction error but is concentrated at depth discontinuities, precisely where blurred transitions back-project to flying pixels. Edge Chamfer Distance decreases in four of five settings, with the largest reductions for LiDM and Wan~$\times$1, confirming that CRISP improves boundary geometry rather than merely increasing local point density. Smooth-region Chamfer Distance decreases by 23\% on average, with only a negligible Wan~$\times$1 exception, confirming that the diffusion decoder does not trade flat-surface accuracy for boundary sharpness. The breakdown thus validates the hypothesis at a spatial granularity unavailable from whole-cloud metrics alone: prior decoders fail selectively at boundaries, and CRISP corrects that.

\paragraph{Encoder-adapted backbones.}
Wan~$\times$1 shows compounding gains: FSVD $-$65\%, FPVD $-$70\%, edge CD $-$60\%, edge F-score $+$100\%, achieving the best perceptual scores of any setting in the study, because encoder adaptation and decoder replacement address complementary
failure modes: the adapted encoder produces a LiDAR-appropriate latent that CRISP can decode without the modality-mismatch artifacts of the convolutional baseline (Fig.~\ref{fig:qual_extra_wan1x}). SVD~$\times$1 improves on all geometric metrics (whole-cloud CD $-$18\%, edge F-score $+$81\%, Fig.~\ref{fig:qual_extra_svd1x}), but FSVD and FPVD regress. We attribute this to reconstruction overfitting: without a perceptual or distributional training signal, CRISP on SVD~$\times$1 minimizes per-sample L1/L2 error at the cost of feature-space realism. Consistently, frozen SVD~$\times$3 already outperforms adapted SVD~$\times$1 on both perceptual metrics. Wan~$\times$1 JSD and EMD similarly regress: we hypothesize that the co-adapted baseline implicitly calibrated its global valid-return density end-to-end, whereas CRISP's support predictor, trained with per-pixel classification losses, does not constrain total point count: a small density miscalibration is sufficient to inflate EMD and JSD without degrading the per-point geometry that FSVD and FPVD reflect.

\paragraph{Plug-and-play world-model decoding.}
Table~\ref{tab:results_world_model} evaluates a zero-shot decoder swap inside a pretrained LiDM world model: the latent generator is frozen, and only the decoder is changed. This is the most demanding test: the world model's latent dynamics were optimised together with the original decoder, yet FSVD still decreases by 15.5\%, FRID by 4.0\%, whole-cloud CD by 2.2\%, and edge CD by 17.5\%. JSD increases marginally, and edge F-score decreases by 20\%, consistent with the latent dynamics having implicitly compensated for the baseline
decoder's oversmoothing: removing that compensation exposes a residual distribution mismatch that fine-tuning the world model alongside CRISP would likely close. The primary geometry improvements nonetheless demonstrate that the decoder bottleneck is actionable in a deployed world model without any retraining (see Fig.~\ref{fig:qual_extra_lidm_wm}).

\paragraph{Inference cost.}
Table~\ref{tab:latency} reports end-to-end throughput. CRISP on Wan2.1 is faster end-to-end than the original pipeline (86\,fps vs.\ 57\,fps) because its decoder (3.84\,ms/frame) is 2.5$\times$ faster than the inherited video VAE decoder. For SVD and LiDM, CRISP's 5-step diffusion decoder runs in 3.9 and 23\,ms/frame respectively, which can be further reduced via distillation~\cite{salimans2022progressive} (see Appendix~\ref{sec:ablation_appendix}, Fig.~\ref{fig:step-ablation}). Thus, despite CRISP's architectural complexity, system cost depends on the inherited backbone: on Wan~$\times$3, it is faster and substantially more accurate than the original decoder.

\begin{table*}[t]
\centering\small
\setlength{\tabcolsep}{4pt}
\caption{Decoder capacity and cost (inherited $\to$ CRISP). Params/FLOPs: batch~1, 5 Euler steps, native range-map resolution per setting. Decoder ms/frame: 8$\times$H200, \texttt{torch.compile}, batch~25, 5 Euler steps, as Table~\ref{tab:latency}.}
\label{tab:capacity}
\begin{tabular}{lccccc}
\toprule
\multirow{2}{*}{Backbone} & \multicolumn{2}{c}{Decoder params} & \multicolumn{2}{c}{GFLOPs} & Decoder ms/frame \\
\cmidrule(lr){2-3}\cmidrule(lr){4-5}
& Inh.\ $\to$ CRISP & Ratio & Inh.\ $\to$ CRISP & Ratio & (Inh.\ $\to$ CRISP, ratio) \\
\midrule
SVD~$\times$3 & 63.6M $\to$ 516.0M & 8.1$\times$ & 376.6 $\to$ 1290.7 & 3.4$\times$ & 1.49 $\to$ 3.92 (2.63$\times$) \\
Wan~$\times$3 & 73.3M $\to$ 516.1M & 7.0$\times$ & 352.6 $\to$ 1291.4 & 3.7$\times$ & 9.71 $\to$ 3.84 (0.40$\times$) \\
LiDM & 8.6M $\to$ 516.0M & 60.0$\times$ & 119.0 $\to$ 2717.2 & 22.8$\times$ & 1.39 $\to$ 23.06 (16.6$\times$) \\
\bottomrule
\end{tabular}
\end{table*}

\paragraph{Decoder capacity.}
CRISP ($\approx$516M parameters) is $8.1\times$, $7.0\times$ and $60.0\times$ larger than the SVD, Wan and LiDM decoders it replaces (Table~\ref{tab:capacity}). A larger decoder could improve reconstruction through added capacity alone, independent of pixel-space diffusion decoding. Two controls show that this extra capacity does not by itself explain the gain. At fixed capacity, the step sweep alone moves SVD~$\times$3 FSVD from $94.2$ at 2 steps to $40.3$ at 5 (Fig.~\ref{fig:step-ablation}); and the gain runs opposite to the added compute, as LiDM receives the most ($60\times$ parameters, $22.8\times$ FLOPs) yet improves least, though partly because its baseline is already closest to its reconstruction ceiling (Appendix~\ref{sec:ablation_appendix}).

\begin{figure}[t]
    \centering
    \includegraphics[width=0.99\linewidth]{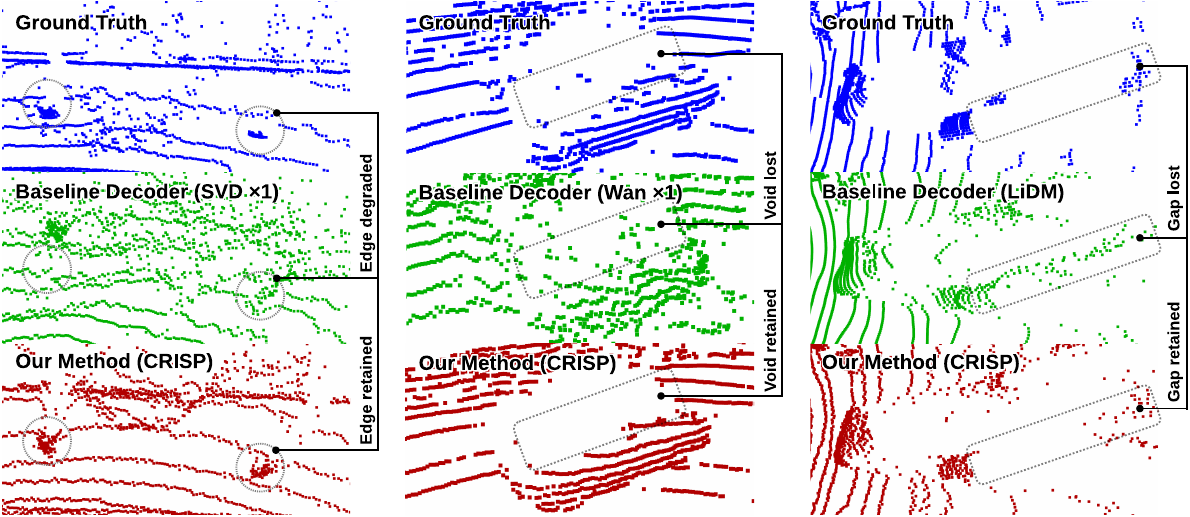}
    \caption{Qualitative LiDAR reconstruction results (colors as in Fig.~\ref{fig:intro_figure}). Columns show, left to right, failure modes CRISP addresses: lost narrow objects, incorrectly filled voids, and flying pixels.}
    \label{fig:qualitative}
\end{figure}

\paragraph{Qualitative analysis.}
Figure~\ref{fig:qualitative} illustrates three distinct decoder failure modes, each suppressed by CRISP: boundary flying pixels (SVD~$\times$1), spurious void infilling (Wan~$\times$1), and depth-gap bridging at foreground-background transitions (LiDM). These correspond directly to the edge and smooth sub-cloud gains in Table~\ref{tab:results_geometry}. Extended per-setting comparisons are in Appendix~\ref{sec:qual_appendix}.

\paragraph{Ablations.}
Appendix~\ref{sec:ablation_appendix} isolates each design choice (sampling steps, early fusion, the mask branch, the loss stack) on SVD~$\times$3, with a per-term leave-one-out analysis (Appendices~\ref{sec:arch:diffloss}--\ref{sec:mask_loss}) and cross-backbone replication on LiDM (Appendix~\ref{sec:lidm_replication}). Early latent fusion is necessary for decoding from pure noise: removing it increases whole-cloud CD by 15\% and FSVD by 21\% (Table~\ref{tab:ablation_dit}). Modeling support with a dedicated mask branch, rather than diffusing over all pixels, improves edge F-score by 23\% and JSD by 35\%, confirming that sparse LiDAR support should be predicted explicitly. The mask ablation further shows the loss stack improves IoU from 0.758 to 0.844, while adding latent conditioning to depth-only input raises converged IoU from 0.630 to 0.788 (Table~\ref{tab:ablation_mask}). Thus, adapter fusion, explicit support branch, and mask supervision each addresses a distinct failure mode rather than merely increasing capacity. The no-mask ablation controls for point-density changes: simultaneous edge-F and edge-CD gains show better boundary placement, not density alone.

\paragraph{Per-component attribution.}
The decoder swap accounts for $68$--$98\%$ of the SVD~$\times$3 gain and the mask branch for the rest, mainly on JSD and edge F-score (Table~\ref{tab:attribution}). The \emph{w/o mask} proxy is not a clean module ablation and beats CRISP only on FPVD ($35.993$ vs.\ $36.782$).

\begin{table}[t]
\centering\small
\setlength{\tabcolsep}{4pt}
\caption{Per-component attribution (SVD~$\times$3, nuScenes, converged): share of the \emph{Inherited}$\to$\emph{CRISP} gain from the decoder swap alone (\emph{w/o mask}) vs.\ the mask branch added on top. \emph{Inherited} from Tables~\ref{tab:results_perceptual}--\ref{tab:results_geometry}; \emph{w/o mask} and \emph{CRISP} from Table~\ref{tab:ablation_dit}.}
\label{tab:attribution}
\begin{tabular}{lrrrrr}
\toprule
Metric & Inherited & w/o mask & CRISP & Decoder swap & Mask branch \\
\midrule
Whole CD $\downarrow$ & 1.331 & 0.660 & 0.639 & 97\% & 3\% \\
FSVD $\downarrow$ & 156.5 & 42.6 & 40.3 & 98\% & 2\% \\
JSD $\downarrow$ & 0.190 & 0.109 & 0.071 & 68\% & 32\% \\
Edge F $\uparrow$ & 0.0952 & 0.193 & 0.238 & 68\% & 32\% \\
\bottomrule
\end{tabular}
\end{table}

\paragraph{Limitations.}
CRISP inherits the range-map assumptions shared by most latent LiDAR pipelines~\cite{ran2024towards,nakashima2024lidar,zyrianov2022learning}: it predicts a single depth channel and does not model remission. Although CRISP suppresses boundary-bridging artifacts and flying pixels, its diffusion-based denoising can still leave residual high-frequency noise in the decoded range maps, especially on smooth surfaces. Since CRISP decodes frames independently, these small artifacts may also appear as thin-structure flicker in multi-frame outputs, a limitation shared with single-frame range-view decoders. The decoder takes 3.9\,ms/frame for SVD and 23\,ms/frame for LiDM (Table~\ref{tab:latency}), so latency remains a limitation for high-throughput downstream perception pipelines. End-to-end, CRISP is faster on Wan~$\times$3 (57$\to$86\,fps) but slower on LiDM (369$\to$41\,fps); distillation~\cite{salimans2022progressive} is future work.

\textbf{Faithfulness to the latent.} A generative decoder could invent boundary geometry the latent does not support. Four measurements indicate that CRISP reads this geometry from the latent instead: edge F-score improves in all five settings and edge CD in four (Table~\ref{tab:results_geometry}); cross-seed edge F-score std is ${\leq}0.0002$ (Appendix~\ref{app:training-protocol}); removing the latent input from the mask predictor drops its IoU from $0.788$ to $0.630$, and removing early latent fusion raises CD from $0.985$ to $1.134$ (Tables~\ref{tab:ablation_mask},~\ref{tab:ablation_dit}); and ungated diffusion over all pixels raises JSD from $0.071$ to $0.109$ (Table~\ref{tab:ablation_dit}). Two residual limits remain: predictor precision is $0.862$ (Table~\ref{tab:ablation_mask}), and in the world model edge F-score falls while edge CD improves (Table~\ref{tab:results_world_model}), which indicates under-prediction rather than fabrication.

\section{Conclusions}
We showed that convolutional VAE decoders, both repurposed and LiDAR-native, are a concrete, correctable bottleneck in latent LiDAR pipelines. Replacing the decoder alone with CRISP recovers sharper geometry from frozen latents, improving edge F-score by 107\% on average and reducing edge CD by 17.5\% in a zero-shot swap inside a frozen LiDM world model, without retraining the latent generator. CRISP therefore offers a plug-in upgrade path for range-map latent LiDAR systems: the generative model can remain unchanged while decoded geometry becomes cleaner and less prone to flying-pixel artifacts. \emph{Future work} includes remission prediction with multi-channel diffusion, density-aware support calibration, and temporal conditioning for flicker-free rollouts.

\begin{ack}
Funding in direct support of this work: salary, compute and research budget provided by BMW AG. Additional revenues related to this work: A.~Ceron is employed by BMW AG as an industrial PhD candidate with TU München; M.~Schmidt and A.~Marcos-Ramiro are full-time employees of BMW AG; S.~Schmidt became a full-time employee of BMW AG during this work.
\end{ack}

\bibliographystyle{plainnat}
\bibliography{references}

\appendix

\section{Extended Related Work}
\label{sec:related_appendix}

\paragraph{Decoder taxonomy mechanism.} LiDM~\cite{ran2024towards} and R2DM~\cite{nakashima2024lidar}, in the LiDARGen lineage~\cite{zyrianov2022learning}, use circular convolutions that wrap around the sensor's $360^\circ$ azimuthal sweep, whereas general-purpose video VAEs such as SVD~\cite{blattmann2023stable} and Wan2.1~\cite{wan2025wan} inherit ordinary image/video padding. Circular padding resolves the azimuthal seam, but any convolutional decoder, regardless of padding, still blends across the \emph{radial} depth discontinuities that cause flying pixels (Sec.~\ref{sec:related}).

\paragraph{Architecture-level structural responses.}
Beyond the post-hoc rectifier of L3DR and the raycasting approach of LidarDM (discussed in Sec.~\ref{sec:related}), several methods address range-view geometric fidelity through representation and architecture choices. TopoLiDM introduces graph-structured latents with topological regularisation to preserve global scene connectivity~\cite{liu2025topolidm}. LidarGRIT explicitly separates range and ray-drop decoding with a vector-quantized transformer, treating valid-return prediction as a distinct problem from depth synthesis~\cite{haghighi2024taming}. RangeFormer identifies projection conflicts and shape deformation as core range-view limitations, motivating a redesigned tokenisation strategy~\cite{kong2023rethinking}. T2LDM targets over-smoothing at the representation level: a self-conditioned representation guidance term aligns the denoiser's intermediate features with real-data representations during training, decoupled at inference, to recover sharper geometric detail without changing the decoder architecture itself~\cite{qu2025t2ldm}. These works improve geometric fidelity through latent representation or tokenisation; none isolates the decoder as a separable stage.

\paragraph{Downstream applications and world modeling.}
Decoder fidelity is consequential in applied settings beyond standalone generation. Adverse-weather augmentation translates clean LiDAR scans to rain, fog, and snow conditions for detector robustness training~\cite{matteazzi2025adverse}, and semantic mask-guided editing synthesises controlled scene variations~\cite{uppur2025range}; in both cases, flying pixels in the decoded output corrupt the resulting training data. U4D further extends world modeling to uncertainty-aware 4D LiDAR prediction~\cite{xu2025u4d}. A growing body of world-model surveys treats latent multimodal representations as the core substrate for autonomous driving systems~\cite{feng2025survey, kong20253d, zeng2026latent, liang2025worldlens}, making decoder quality increasingly consequential as generation is embedded in closed-loop evaluation and planning pipelines.

\paragraph{Support prediction in LiDAR generation.}
Explicit support prediction is a standard component of LiDAR generation. Prior work models ray-dropping as learned binary classification with U-Net-style architectures~\cite{ronneberger2015u, manivasagam2020lidarsim, zyrianov2025lidardm}, and support prediction has also appeared in latent LiDAR pipelines~\cite{zheng2024lidar4d}.

\paragraph{Pixel-space diffusion for depth.}
The pixel-space diffusion lineage relevant to CRISP begins with Marigold, which showed that pretrained diffusion priors can be repurposed for dense monocular depth estimation through finetuning on synthetic data~\cite{ke2024repurposing}. This established that generative models carry geometric priors useful beyond image synthesis, but the decoder in that setting operates on a dense, support-complete depth map and the 3D consequences of decoder-induced smoothing were not the focus. Pixel-Perfect Depth carries this further with a cascade of diffusion transformers~\cite{peebles2023scalable} conditioned on a clean RGB image, showing that even ground-truth-quality depth is degraded by VAE decoding~\cite{xu2025pixel}; CRISP adapts this cascade-DiT idea to a regime with no clean guidance image and genuine missing returns (Sec.~\ref{sec:related}). 

Concretely, CRISP differs from Pixel-Perfect Depth on three axes: removing CRISP's early latent injection, which replaces Pixel-Perfect Depth's persistent RGB conditioning, costs whole-cloud CD $0.985\!\to\!1.134$ and FSVD $98.6\!\to\!119.6$; diffusing over all pixels instead of gating through CRISP's predicted support mask (needed because LiDAR range maps, unlike dense monocular depth, have genuine missing returns) costs JSD $0.071\!\to\!0.109$ and edge F-score $0.238\!\to\!0.193$ (both Table~\ref{tab:ablation_dit}); and unlike Pixel-Perfect Depth's single continuous depth output, CRISP composes a depth map with a predicted validity mask, $\hat{y} = \hat{m} \odot \hat{y}_d - (1-\hat{m})$. Two recent lines replace convolutional decoders with diffusion decoders for natural images rather than depth: PiD unifies latent decoding and super-resolution into a single diffusion pass, distilled to a handful of steps for speed~\cite{lu2026pid}, and $\epsilon$-VAE reframes VAE decoding itself as iterative denoising~\cite{zhao2025epsilon}; both are optimized for perceptual realism rather than per-pixel faithfulness to the latent (Sec.~\ref{sec:related}). LiDAR range maps, by contrast, are geometric measurements feeding safety-relevant perception, where per-frame faithfulness rather than distributional typicality is required, which is why CRISP's depth branch is a validity-masked regression against ground truth rather than a free sampler. RAE moves in the opposite direction, replacing the VAE encoder with a frozen pretrained representation encoder (e.g.\ DINO, SigLIP, or MAE) paired with a trained decoder, and argues that the VAE's low-dimensional, weakly semantic latent limits diffusion transformers~\cite{zheng2025diffusion}. CRISP instead keeps the encoder and the pretrained latent generator entirely frozen, which is what allows a plug-in decoder swap inside an already-trained LiDM world model (Table~\ref{tab:results_world_model}).

\section{Architectural Details}
\label{sec:arch_appendix}

\subsection{DiT Decoder}
\label{sec:arch:dit}

\paragraph{Model scale.}
The DiT decoder $\mathcal{D}_{\theta}$ has $K{=}24$ transformer blocks, hidden dimension $C_d{=}1024$, and $16$ attention heads. Each block uses an MLP with ratio~$4.0$ and tanh-approximated GELU activation~\cite{hendrycks2016gaussian}. Attention uses QK normalisation for training stability~\cite{henry2020query}. All weight matrices are Xavier-initialised; adaLN modulation projection weights are zero-initialised so every block is an identity map at the start of training~\cite{peebles2023scalable}.

\paragraph{Coarse-to-fine cascade.}
The noisy input $x_t$ is tokenised into $p{\times}p = 16{\times}16$ patches, giving $T_0 = \frac{H}{16}\cdot\frac{W}{16}$ tokens. Blocks $0$-$(K/2{-}1)$ process these at the coarse scale with 2D RoPE at resolution $(H/16, W/16)$~\cite{su2024roformer}. At the mid-fusion step, the $T_0$ token sequence is pixel-shuffled to $4T_0 = \frac{H}{8}\cdot\frac{W}{8}$ tokens (a reshape-einsum operation equivalent to spatial unfolding), and blocks $K/2$-$(K{-}1)$ refine at this finer scale with RoPE at $(H/8, W/8)$. The coarse-first progression lets global layout be established at low token count before local geometry is refined at higher resolution, which matters because the decoder starts from pure Gaussian noise with no auxiliary observation.

\paragraph{Latent fusion asymmetry.}
Adapter tokens $\tau$ are L2-normalised before both injections. Early fusion is applied to the token stream \emph{before} block~$0$ processes its input, using a lightweight two-layer MLP ($2C_d \!\to\! C_d \!\to\! C_d$, SiLU): the global layout signal must enter cheaply so it is available from the first block. Mid-level fusion (block $K/2{-}1$) uses a wider three-layer MLP ($2C_d \!\to\! 4C_d \!\to\! 4C_d \!\to\! 4C_d$): this projection simultaneously fuses geometry-aware context and provides the spatial token counts required for the pixel-shuffle upsample, so additional capacity is justified.

\subsection{Support Mask Predictor}
\label{sec:arch:mask}

$\mathcal{M}_{\phi}$ is a U-Net with three encoder levels and base channel width $C_m{=}128$ ($C_m{\to}2C_m{\to}4C_m$)~\cite{ronneberger2015u}. Standard components (skip connections, average-pool downsampling, bilinear upsampling, GroupNorm~\cite{wu2018group}, Xavier init) follow established practice.

\paragraph{Dual-branch conditioning.}
A latent branch processes $z$ through a two-layer convolutional block (Conv-GroupNorm-GELU $\times 2$) and upsamples it bilinearly to the range-map resolution. A separate depth-hint branch maps the single decoded depth channel to $C_m/2{=}64$ features through a shallow convolutional block. The two branches are concatenated ($C_m + C_m/2$) and fused before the U-Net encoder stack, giving the predictor access to both semantic layout (from the latent) and high-frequency geometric structure (from the depth hint).

\subsection{Geometry-Aware Diffusion Objective}
\label{sec:arch:diffloss}

\paragraph{Valid-region velocity loss.}
The base supervision signal is a plain validity-masked regression on the velocity target ($\hat{v}$, $v^\star$ as in Eq.~\ref{eq:diffusion}):
\begin{equation}
  \mathcal{L}_{\mathrm{vel,raw}}
  = \Bigl(\sum\nolimits_{ij} m_{ij}\,\ell(\hat{v}_{ij},\,v^\star_{ij})\Bigr)
  \Big/ \Bigl(\sum\nolimits_{ij} m_{ij}\Bigr),
  \label{eq:vel_raw}
\end{equation}
where $\ell$ is $\ell_2$ until validation depth metrics plateau, then $\ell_1$ for fine-grained convergence. Every valid pixel contributes equally.

\paragraph{Directional finite-difference losses.} Range maps are anisotropic: horizontal neighbors lie along the same LiDAR beam sweep, while vertical neighbors correspond to different elevation beams at the same azimuth. We therefore regularize local structure along both axes. First-order terms preserve local slopes, while second-order terms suppress high-frequency waviness introduced by denoising. Letting $\delta_x v_{ij} = v_{i,j+1}-v_{i,j}$, $\delta_{xx} v_{ij} = v_{i,j+1}-2v_{i,j}+v_{i,j-1}$, and $\delta_y$, $\delta_{yy}$ the analogous vertical operators, all four terms share the form:
\begin{equation}
  \mathcal{L}_{d\alpha}
  = \frac{1}{|\Omega_\alpha|}\sum_{\Omega_\alpha}
    \bigl|\delta_\alpha \hat{v}_{ij} - \delta_\alpha v^\star_{ij}\bigr|,
  \quad \alpha \in \{x,\, xx,\, y,\, yy\},
  \label{eq:der_terms}
\end{equation}
where $\Omega_\alpha$ contains pixels whose neighbors involved in $\delta_\alpha$ all satisfy $m=1$. The combined loss is:
\begin{equation}
  \mathcal{L}_{\mathrm{der}}
  =
  \lambda_{dx} \mathcal{L}_{dx}
  + \lambda_{d2x} \mathcal{L}_{d2x}
  + \lambda_{dy} \mathcal{L}_{dy}
  + \lambda_{d2y} \mathcal{L}_{d2y},
  \qquad \lambda_{dx},\lambda_{d2x},\lambda_{dy},\lambda_{d2y} \geq 0.
  \label{eq:der_loss}
\end{equation}

\paragraph{Low-noise reconstruction term.} As a fine-tuning signal, $\mathcal{L}_{x0}$ penalizes the denoised estimate $\hat{y}_d$ against the target range map, with a linear timestep weight that emphasizes low-noise steps:
\begin{equation}
  \mathcal{L}_{x0}
  =
  \frac{1}{\sum_{ij}m_{ij}}
  \sum_{ij}
  \Bigl(1-\tfrac{t}{T}\Bigr)\,m_{ij}\,
  \bigl|\hat{y}_{d,ij} - \tilde{y}_{ij}\bigr|.
  \label{eq:x0_loss}
\end{equation}

\paragraph{Boundary-aware velocity loss (weighted form).}
The boundary-aware velocity loss $\mathcal{L}_{\mathrm{vel}}$ referenced in the main text replaces $m$ with the importance weight $w$ from
Eq.~\ref{eq:importance}:
\begin{equation}
  \mathcal{L}_{\mathrm{vel}}
  = \Bigl(\sum\nolimits_{ij} w_{ij}\,\ell(\hat{v}_{ij},\,v^\star_{ij})\Bigr)
  \Big/ \Bigl(\sum\nolimits_{ij} w_{ij}\Bigr).
  \label{eq:vel_weighted}
\end{equation}
The structure signal uses $G_\sigma$ with kernel $5{\times}5$, $\sigma{=}2.0$, and cap $\lambda_{\max}{=}1.5$. Loss weights for all frozen and encoder-adapted training stages are listed in Table~\ref{tab:loss-weights}.

\paragraph{Depth-branch leave-one-out.} Table~\ref{tab:depth-loo} reports validation L1 ($\downarrow$) when each depth term is removed (for $\mathcal{L}_{\mathrm{vel}}$, its importance weighting is set uniform, $w\!\to\!m$) from an otherwise-matched reference run, on SVD~$\times$3/nuScenes and, separately, on the frozen LiDAR-native LiDM backbone/KITTI-family. Every surviving term is load-bearing on at least one dataset, and the two disagree about which: the $\ell_1$ pixel penalty $\mathcal{L}_{x0}$ matters on nuScenes but far less on KITTI-family, while the finite-difference term $\mathcal{L}_{\mathrm{der}}$ shows the reverse, consistent with sharper radial discontinuities at 64 beams. Dropping either would silently specialise the objective to one beam count. $\mathcal{L}_{\mathrm{grad}}$ is the only term small on both datasets and is the one dropped from the four-term reduced depth objective (Table~\ref{tab:loss-provenance}).
\begin{table}[h]
\centering\small
\caption{Depth-branch leave-one-out: validation L1 ($\downarrow$) when each term is removed from the reference objective (for $\mathcal{L}_{\mathrm{vel}}$, its importance weighting is set uniform), on SVD~$\times$3/nuScenes and on the frozen LiDM backbone/KITTI-family; percentages are the change against the matched reference.}
\label{tab:depth-loo}
\begin{tabular}{lcc}
\toprule
Removed term & nuScenes val L1 & KITTI-family val L1 \\
\midrule
-- (reference) & 0.06052 & 0.0218291 \\
Importance weighting ($w\!\to\!m$ in $\mathcal{L}_{\mathrm{vel}}$) & 0.07578 ($+$25.2\%) & 0.0231271 ($+$5.9\%) \\
$\ell_1$ pixel penalty ($\mathcal{L}_{x0}$) & 0.06540 ($+$8.1\%) & 0.0218987 ($+$0.3\%) \\
Finite differences ($\mathcal{L}_{\mathrm{der}}$) & 0.06291 ($+$3.9\%) & 0.0235116 ($+$7.7\%) \\
Multi-scale gradient ($\mathcal{L}_{\mathrm{grad}}$) & 0.06228 ($+$2.9\%) & 0.0218484 ($+$0.1\%) \\
\bottomrule
\end{tabular}
\end{table}

\subsection{Mask loss weights.}
\label{sec:mask_loss}

The mask objective (Eq.~\ref{eq:mask_loss}) combines six terms with the following weights:
\begin{table}[h]
\centering\small
\caption{Mask loss sub-term weights (Eq.~\ref{eq:mask_loss}).}
\label{tab:mask-weights}
\begin{tabular}{lc}
\toprule
Term & $\lambda$ \\
\midrule
Focal BCE ($\gamma\!=\!2$, $\alpha\!=\!0.4$) & 1.0 \\
Soft Dice                                      & 1.0 \\
Edge (1st-order)                               & 0.5 \\
Laplacian (2nd-order)                          & 1.5 \\
Hard-pixel mining                              & 0.5 \\
Deep supervision                               & 0.5 \\
\bottomrule
\end{tabular}
\end{table}

\paragraph{Mask-branch leave-one-out and minimal subset.} Table~\ref{tab:mask-loo} reports the paired change in IoU when each term is removed from the full six-term objective against a matched, same-seed reference (SVD~$\times$3/nuScenes; the frozen LiDM backbone/KITTI-family, reference IoU 0.9888). Focal BCE and soft Dice separate the two datasets: load-bearing on nuScenes, but next to unresolvable on KITTI-family, where support prediction on the LiDAR-native backbone is already near ceiling. Table~\ref{tab:mask-minimal} instead trains only a subset of terms (LiDM backbone/KITTI-family, 10 epochs, seed 23, batch 16; weights Focal~1.0, Dice~1.0, Laplacian~1.5, the rest zeroed). Three terms, Focal~+~Dice~+~Laplacian, recover 96.7\% of the full objective's IoU gain over focal-only supervision, and outperform the Focal~+~Dice~+~Edge combination an additive sweep would select instead (cf.\ Table~\ref{tab:ablation_mask}): an ordering artifact, since Edge enters first in that cumulative ablation and absorbs the shared boundary term. The minimal mask objective is therefore Focal~+~Dice~+~Laplacian, three of six (Sec.~\ref{sec:method:maskloss}).
\begin{table}[h]
\centering\small
\caption{Mask-branch leave-one-out: paired change in IoU vs.\ a matched, same-seed full-objective reference when each term is removed.}
\label{tab:mask-loo}
\begin{tabular}{lcc}
\toprule
Removed term & nuScenes $\Delta$IoU & KITTI-family $\Delta$IoU \\
\midrule
Laplacian & $-$0.00245 & $-$0.00110 \\
Soft Dice & $-$0.00206 & $-$0.00010 \\
Focal BCE & $-$0.00172 & $-$0.00010 \\
Hard-pixel mining & $-$0.00054 & $\approx$0 \\
Deep supervision & $-$0.00034 & $\approx$0 \\
Edge & $-$0.00033 & $-$0.00020 \\
\bottomrule
\end{tabular}
\end{table}
\begin{table}[h]
\centering\small
\caption{Mask-branch minimal subset (LiDM backbone, KITTI-family, 10 epochs): converged IoU when zeroing all but the listed terms, and the resulting share of the full objective's IoU gain over focal-only supervision retained.}
\label{tab:mask-minimal}
\begin{tabular}{lcc}
\toprule
Mask objective & IoU & Gain retained \\
\midrule
Focal only & 0.97418 & -- \\
Focal + Dice + Edge & 0.98761 & 91.9\% \\
Focal + Dice + Laplacian & 0.98831 & 96.7\% \\
Full six-term & 0.98879 & 100\% \\
\bottomrule
\end{tabular}
\end{table}

\subsection{Objective Design: Provenance, Balancing, and Compute Cost}
\label{sec:loss_justification}

Table~\ref{tab:loss-provenance} lists, for every term in Eqs.~\ref{eq:depth_loss} and~\ref{eq:mask_loss}, the failure mode it is admitted for, the literature precedent for the operator, its weight (Stage~F2, Table~\ref{tab:loss-weights}), and whether it survives in the reduced objective (Tables~\ref{tab:depth-loo}-\ref{tab:mask-minimal}).
\begin{table}[h]
\centering\footnotesize
\setlength{\tabcolsep}{3pt}
\caption{Loss-term justification: failure mode, operator provenance, weight, and reduced-objective status for every term in $\mathcal{L}_{\mathrm{depth}}$ (Eq.~\ref{eq:depth_loss}) and $\mathcal{L}_{\mathrm{mask}}$ (Eq.~\ref{eq:mask_loss}).}
\label{tab:loss-provenance}
\begin{tabular}{lp{4.8cm}p{4.8cm}cc}
\toprule
Term & Failure mode & Provenance & Weight & Reduced \\
\midrule
\multicolumn{5}{l}{\textbf{Depth branch (Eq.~\ref{eq:depth_loss})}} \\
$\mathcal{L}_{\mathrm{vel}}$ & region: boundary-aware velocity regression on valid pixels & -- & 1.0 & yes \\
$\mathcal{L}_{\mathrm{vel,raw}}$ & stable unweighted baseline; normalises $\mathcal{L}_{\mathrm{vel}}$'s per-edge gradient share & structural, no operator citation & 1.0 & yes \\
$\mathcal{L}_{\mathrm{der}}$ & anisotropic per-ring radial bias invisible per-pixel & per-axis derivative matching (cf.\ LiDM gradient loss~\cite{ran2024towards}) & Table~\ref{tab:loss-weights} & yes \\
$\mathcal{L}_{x0}$ & low-noise accuracy of $\hat{y}_d$ (fine-tuning signal) & timestep-weighted diffusion loss~\cite{salimans2022progressive, song2020score} & 0.5 & yes \\
$\mathcal{L}_{\mathrm{grad}}$ & low-frequency range-gradient residual & LiDM multi-scale gradient loss~\cite{ran2024towards} & 0.5 & no \\
\midrule
\multicolumn{5}{l}{\textbf{Mask branch (Eq.~\ref{eq:mask_loss})}} \\
$\mathcal{L}_{\mathrm{focal}}$ & region: valid/void class imbalance & focal loss for dense detection~\cite{lin2017focal} & 1.0 & yes \\
$\mathcal{L}_{\mathrm{dice}}$ & scarce thin/sparse valid-return recall & overlap-based segmentation loss~\cite{milletari2016v} & 1.0 & yes \\
$\mathcal{L}_{\mathrm{lap}}$ & isolated single-pixel ray-drop speckle & depth-completion smoothness~\cite{ma2019self,zhong2019deep} & 1.5 & yes \\
$\mathcal{L}_{\mathrm{edge}}$ & boundary misalignment (first-order instance of $\mathcal{L}_{\mathrm{lap}}$'s prior) & boundary-consistency loss~\cite{huang2019indoor} & 0.5 & no \\
$\mathcal{L}_{\mathrm{hard}}$ & under-trained near-boundary pixels & online hard-example mining~\cite{shrivastava2016training} & 0.5 & no \\
$\mathcal{L}_{\mathrm{ds}}$ & weak gradient signal at shallow layers & deep supervision~\cite{lee2015deeply,qin2019basnet} & 0.5 & no \\
\bottomrule
\end{tabular}
\end{table}

\paragraph{Term count.} In the frozen setting the objective comprises eleven loss terms: five in $\mathcal{L}_{\mathrm{depth}}$ and six in $\mathcal{L}_{\mathrm{mask}}$; encoder-adapted variants add the KL term of Eq.~\ref{eq:full_loss_adapt}. A count of fourteen arises only if $\mathcal{L}_{\mathrm{der}}$ (Eq.~\ref{eq:der_loss}) is unpacked into its four $(\text{axis},\text{order})$ components, $\mathcal{L}_{dx}$, $\mathcal{L}_{d2x}$, $\mathcal{L}_{dy}$, $\mathcal{L}_{d2y}$ (Eq.~\ref{eq:der_terms}), and each counted separately. They are computed separately because the range map is a 2D grid with two physically distinct axes (horizontal beam sweep, vertical elevation) and because first- and second-order differences capture different structure (slope vs.\ curvature), not because each is an independently motivated objective.

\paragraph{Balancing procedure.} Per-term weights (Table~\ref{tab:loss-weights}) are set once per training stage: every term is scaled to an equal initial magnitude and then lightly adjusted with a simplified, static form of GradNorm~\cite{chen2018gradnorm} rather than its full dynamic procedure. The same resulting weight vector is used for every backbone (SVD, Wan2.1, LiDM) and both beam counts (32-beam nuScenes, 64-beam KITTI-family), with no per-setting retuning. Apart from $\mathcal{L}_{x0}$, $\mathcal{L}_{\mathrm{grad}}$ and the mask branch, which are switched on at Stage~F2, the derivative-loss weights are the only ones that change within a run, up-weighted 8-17$\times$ between Stage~F1 and Stage~F2 (Table~\ref{tab:loss-weights}); this did not destabilise training, with all twenty unsmoothed validation checkpoints on the matched reference run improving monotonically. The one resulting sensitivity, gradient clipping on 64-beam KITTI-family, is discussed in Appendix~\ref{app:training-protocol}.

\paragraph{Compute and memory cost.} Table~\ref{tab:loss-compute} measures training step time and peak memory as loss terms are added, holding the network, data, and checkpoints fixed so only the objective changes (one H200, BF16, SVD~$\times$3/nuScenes, 24-block DiT, 25-frame clip; median of 100 steps after 50 warm-up steps, five interleaved runs per configuration; between-run spread 1.09-2.18\,ms). The measurable cost is the second branch (the mask predictor's forward, loss, and backward pass), not the number of terms: going from the minimal two-branch configuration ($\mathcal{L}_{\mathrm{vel}}$, $\mathcal{L}_{\mathrm{vel,raw}}$, $\mathcal{L}_{\mathrm{focal}}$) to the full eleven-term objective adds 7.81\,ms, 3.0\% of a step, at effectively identical peak memory (34.397 vs.\ 34.394\,GiB), because the added terms are elementwise or finite-difference operations on single-channel maps with no parameters and no extra forward pass. Individual terms cost at or below the between-run spread, so we do not report them separately. The seven-term reduced objective (Tables~\ref{tab:depth-loo}-\ref{tab:mask-minimal}) is motivated by overlap between terms, not by this already-small compute: it saves only 5.36\,ms, 2\% of a step, versus the full objective. At inference no loss is computed, but the mask branch itself still runs: 295.49\,ms per clip vs.\ 252.46\,ms with the branch stubbed out, i.e.\ 43.03\,ms, or 14.6\% of decoder latency (batch~1, 25-frame clip, 5 Euler steps).
\begin{table}[h]
\centering\small
\caption{Training step time and peak memory as loss terms are added, network/data/checkpoints held fixed (one H200, BF16, SVD~$\times$3/nuScenes, median of 100 steps).}
\label{tab:loss-compute}
\begin{tabular}{lccrr}
\toprule
Objective & Branches & Terms & ms/step$\downarrow$ & Peak GiB$\downarrow$ \\
\midrule
Depth only ($\mathcal{L}_{\mathrm{vel}}\!+\!\mathcal{L}_{\mathrm{vel,raw}}$) & Depth & 2 & 165.11 & 19.878 \\
$+\,\mathcal{L}_{\mathrm{focal}}$ & Depth + Mask & 3 & 263.23 & 34.397 \\
Reduced set & Depth + Mask & 7 & 265.68 & 34.393 \\
Full objective & Depth + Mask & 11 & 271.04 & 34.394 \\
\bottomrule
\end{tabular}
\end{table}

\section{Additional Experimental Details}
\label{app:experimental}

\subsection{Data Preparation}
\label{app:data-preprocessing}
\paragraph{KITTI family (KITTI-360, SemanticKITTI).}
Raw 64-beam point clouds are projected onto $64\!\times\!1024$ spherical range images following the LiDM projection protocol (vertical FoV $[-25^\circ, +3^\circ]$, azimuthal range $360^\circ$). Valid depths lie in $[1.0,\,56.0]$ m; each depth value $d$ is log-normalised and mapped to $[-1, 1]$: 
\begin{equation}
  \hat{d} = \frac{2\,\log_2(d + 1)}{\delta} - 1,
  \qquad \delta = 5.84,
  \label{eq:kitti-norm}
\end{equation}
with empty bins assigned $\hat{d}=-1$. The training split provided by LiDM contains 4,376 25-frame LiDAR clips. We apply the LiDM augmentation protocol 5× (identity, vertical flip, horizontal flip, point dropout, and horizontal shift), yielding 21,880 augmented 25-frame clips per epoch. Although clips define the sampling unit, all frames are encoded and decoded independently. The validation split is used as provided; no test-set frames are used for any model-selection decision.

\paragraph{nuScenes.}
Each scan is projected onto a $32\!\times\!1024$ beam-indexed range image (one row per laser ring), with valid depths in $[2.0,\,45.0]$ m, log-normalised to $[-1, 1]$:
\begin{equation}
  \hat{d} = \frac{2\bigl(\log_2(d+1) - \log_2(d_{\min}+1)\bigr)}
                 {\log_2(d_{\max}+1) - \log_2(d_{\min}+1)} - 1,
  \label{eq:nus-norm}
\end{equation}
with $d_{\min}=2.0$ m and $d_{\max}=45.0$ m; empty bins are set to $-1$. We follow the VISTA~\cite{gao2024vista} data partition: the official VISTA \emph{training} set is split into a training portion of 19,088 25-frame LiDAR clips and a held-out validation subset; the VISTA test set is reserved exclusively for evaluation. Although clips provide the sampling unit, all frames are encoded and decoded independently, without temporal compression.

\subsection{Frozen Backbones.}
\label{app:frozen-backbones}
For the frozen-encoder ($\times$3) variants, neither the SVD nor the Wan2.1 VAE encoder receives any parameter updates. The single-channel range image is replicated across all three input channels before encoding ($\hat{d}_{\text{rgb}} = \hat{d} \cdot \mathbf{1}_3$). For the inherited-decoder baseline, the original VAE decoder outputs a three-channel reconstruction, which we convert back to a single-channel range map by averaging the three decoded channels. For CRISP, the original decoder is replaced by the CRISP DiT decoder, which directly predicts a single-channel range map from the same frozen latent. No other modifications are made to the backbone.

\subsection{Encoder Adaptation.}
\label{app:encoder-adapt}
Both the SVD and Wan2.1 VAEs are pretrained on 3-channel RGB video. For the \emph{adapted baseline} setting, we follow the encoder-adaptation pipeline proposed in UniDriveDreamer~\cite{zhao2026unidrivedreamer} without modification: the VAE is converted to single-channel LiDAR input/output by weight surgery and then fine-tuned with the same reconstruction objective. This baseline answers whether CRISP remains useful after the inherited VAE itself has been adapted to LiDAR. 

Both adapted baseline VAEs use the UniDriveDreamer-style objective:
\[
  \mathcal{L}_{\text{enc}}
  = \lambda_{\text{recon}}\mathcal{L}_1
  + \lambda_{\text{kl}}\mathcal{L}_{\text{KL}}
  + \lambda_{\text{lpips}}\mathcal{L}_{\text{LPIPS}},
\]
where LPIPS uses a frozen VGG backbone~\cite{zhang2018unreasonable}. Since VGG-LPIPS expects three-channel input, we replicate the single-channel normalised LiDAR range map across RGB channels before computing LPIPS. This objective is used only to produce the adapted VAE baselines; CRISP-adapted variants instead train the encoder jointly with the CRISP depth and mask objectives plus the KL regulariser in Eq.~\ref{eq:full_loss_adapt}.

\paragraph{SVD~$\times$1.}
The SVD VAE is evaluated in single-frame mode: although its video decoder contains temporal-mixing modules, we pass \texttt{timesteps=1}, so these modules reduce to per-frame operations and do not introduce temporal-compression artifacts. No layer-wise freeze tiers are needed: all parameters are trained at a uniform learning rate $\eta_{\text{vae}} = 5\!\times\!10^{-5}$. Before any gradient updates, the 3-channel I/O weights are averaged into 1-channel initialisations: \texttt{encoder.conv\_in} ($[128,3,3,3]\!\to\![128,1,3,3]$, averaged over the input-channel axis) and \texttt{decoder.conv\_out} ($[3,128,3,3]\!\to\![1,128,3,3]$, averaged over the output-channel axis); biases and temporal-mixing parameters are kept from the pretrained checkpoint. This preserves pretrained filter magnitudes while yielding a valid 1-channel starting point. Loss weights: $(\lambda_{\text{recon}},\lambda_{\text{kl}},\lambda_{\text{lpips}}) = (1.0,\,10^{-5},\,0.3)$.

\paragraph{Wan2.1~$\times$1.}
Wan2.1 introduces four purely-temporal convolutions (\texttt{encoder.downsamples.\allowbreak\{5,8\}.\allowbreak time\_conv}, \texttt{decoder.upsamples.\allowbreak\{3,7\}.\allowbreak time\_conv}) that implement $\times4$ temporal compression.  Updating these layers (even at extremely low learning rates) causes the model to collapse within a handful of iterations.  We therefore \emph{freeze all four temporal convolutions entirely} and apply a two-tier learning rate: the four I/O convolutions (\texttt{encoder.conv1}, \texttt{encoder.head}, \texttt{decoder.conv1}, \texttt{decoder.head}) and the two latent projection layers (\texttt{conv1}, \texttt{conv2}), which undergo the largest domain shift, are trained at $\eta_{\text{vae}} = 5\!\times\!10^{-5}$; all remaining interior ResBlocks at $0.1\,\eta_{\text{vae}}$.  Loss weights: $(\lambda_{\text{recon}},\lambda_{\text{kl}},\lambda_{\text{lpips}}) = (1.0,\,10^{-2},\,0.3)$.  The larger $\lambda_{\text{kl}}$ compared to SVD reflects the temporal architecture's greater tolerance for KL regularisation after the I/O surgery.

\subsection{Temporal Compression Bypass.}
\label{app:temporal-bypass}
Wan2.1 applies $\times4$ temporal compression, under which only the first frame in the encoded clip is reconstructed best, while subsequent frames progressively degrade. Table~\ref{tab:temporal-ablation} quantifies this for Wan~$\times$3 and Wan~$\times$1 over 25 frames: at $t\!=\!0$, CD is $0.770$ / $0.309$ for $\times$3 / $\times$1; by $t\!=\!5$ it reaches $0.806$ / $0.377$ ($\Delta\text{CD}=+0.036$ / $+0.068$).  To eliminate this artifact in the main comparison, we evaluate both the inherited Wan decoder and CRISP independently on each frame, feeding 25 separate single-frame latents rather than a single temporally compressed video tensor.
\begin{table}[h]
\centering
\caption{Per-frame CD ($\downarrow$) and F-score@0.5m ($\uparrow$) for Wan~$\times$3 (left) and Wan~$\times$1 (right) under joint temporal encoding. Frame $t\!=\!0$ is systematically the best; the degradation motivates per-frame inference. \textbf{Bold} = best per column.}
\label{tab:temporal-ablation}
\small
\begin{tabular}{c cc cc}
\toprule
 & \multicolumn{2}{c}{Wan~$\times$3} & \multicolumn{2}{c}{Wan~$\times$1} \\
$t$ & CD$\downarrow$ & F-score@0.5m$\uparrow$ & CD$\downarrow$ & F-score@0.5m$\uparrow$ \\
\midrule
 0 & \textbf{0.770} & \textbf{0.732} & \textbf{0.309} & \textbf{0.867} \\
 1 & 0.792 & 0.718 & 0.336 & 0.862 \\
 2 & 0.805 & 0.712 & 0.350 & 0.859 \\
 3 & 0.794 & 0.721 & 0.333 & 0.862 \\
 4 & 0.804 & 0.716 & 0.351 & 0.858 \\
 5 & 0.806 & 0.710 & 0.377 & 0.854 \\
\multicolumn{1}{c}{\vdots} & \multicolumn{4}{c}{\vdots} \\
Mean & 0.795 & 0.717 & 0.353 & 0.857 \\
\bottomrule
\end{tabular}
\end{table}

\section{Training Protocol}
\label{app:training-protocol}
The training protocol differs between frozen-encoder ($\times$3) and encoder-adapted ($\times$1) variants.

\paragraph{Frozen-encoder variants ($\times$3): two stages.}
Frozen-encoder variants use two stages. In Stage~F1, the encoder is fixed and only the DiT decoder is trained with an L$_2$ velocity objective; the mask predictor is absent. In Stage~F2, the encoder remains fixed, the objective switches from L$_2$ to L$_1$, gradient and derivative losses are up-weighted, and the support mask predictor is introduced and trained jointly with the DiT. We transition from Stage~F1 to Stage~F2 after $\approx$12~h on 8$\times$A100 GPUs, when validation reconstruction loss plateaus.

\paragraph{Encoder-adapted variants ($\times$1): three stages.}
Encoder-adapted variants use three stages. In Stage~A1, the encoder and DiT decoder are trained jointly with an L$_2$ velocity objective and KL regularisation; the mask predictor is absent. In Stage~A2, the mask predictor is added while the encoder remains trainable and the depth branch still uses the L$_2$ objective. This lets the mask branch adapt to the latent space without prematurely dominating the initial encoder/DiT training. In Stage~A3, the encoder is frozen, the objective switches from L$_2$ to L$_1$, gradient and derivative losses are up-weighted, and the DiT and mask predictor are jointly fine-tuned on the fixed latent manifold.

\paragraph{Runtime and compute.}
Each CRISP training run takes approximately 36~h on 8$\times$A100-80GB GPUs: about 24~h for the DiT-only stage and an additional 12~h after introducing the support mask branch. This corresponds to roughly 288 GPU-hours per full CRISP run. Runtime varies slightly across backbones and excludes baseline VAE adaptation and ablations.

\begin{table}[h]
\centering
\caption{Training stages and loss weights. Frozen-encoder variants ($\times$3) use Stages~F1-F2; encoder-adapted variants ($\times$1) use Stages~A1-A3. ``--'' denotes an inactive branch or loss.}
\label{tab:loss-weights}
\small
\setlength{\tabcolsep}{4pt}
\begin{tabular}{lccccc}
\toprule
\multirow{2}{*}{Loss / setting}
& \multicolumn{2}{c}{Frozen encoder ($\times$3)}
& \multicolumn{3}{c}{Encoder-adapted ($\times$1)} \\
\cmidrule(lr){2-3}\cmidrule(lr){4-6}
& F1 & F2 & A1 & A2 & A3 \\
\midrule
Encoder trainable
& no & no & yes & yes & no \\
Mask branch
& -- & yes & -- & yes & yes \\
Velocity loss
& L$_2$ & L$_1$ & L$_2$ & L$_2$ & L$_1$ \\
\midrule
$\lambda_{x0}$ 
& 0.0 & 0.5 & 0.0 & 0.0 & 0.5 \\
$\lambda_{\mathrm{grad}}$
& 0.0 & 0.5 & 0.0 & 0.0 & 0.5 \\
$\lambda_{dx}$
& 0.25 & 2.0 & 0.25 & 0.25 & 2.0 \\
$\lambda_{d2x}$
& 0.25 & 2.0 & 0.25 & 0.25 & 2.0 \\
$\lambda_{dy}$
& 0.12 & 2.0 & 0.12 & 0.12 & 2.0 \\
$\lambda_{d2y}$
& 0.12 & 2.0 & 0.12 & 0.12 & 2.0 \\
$\lambda_{\mathrm{KL}}$
& -- & -- & $5{\times}10^{-6}$ & $5{\times}10^{-6}$ & -- \\
$\lambda_{\mathrm{mask}}$
& -- & 1.2 & -- & 1.2 & 1.2 \\
\bottomrule
\end{tabular}
\end{table}

Importance weights emphasise geometrically complex regions using the valid-only local structure signal from Eqs.~\ref{eq:valid_blur}-\ref{eq:importance}, with a $5{\times}5$ Gaussian kernel, $\sigma=2.0$, and cap $\lambda_{\max}=1.5$. During Stage~A2, mask gradients are scaled by $0.1$ when back-propagated through the shared encoder, preventing mask-classification gradients from corrupting the depth latent representation.

\paragraph{Optimizer and Regularization.}
All models use AdamW.  Learning rates: $\eta_{\text{DiT}} = \eta_{\text{mask}} = 10^{-4}$.  For frozen-encoder ($\times$3) variants, $\eta_{\text{enc}} = 0$ throughout.  For encoder-adapted ($\times$1) variants, $\eta_{\text{enc}} = 5\!\times\!10^{-5}$ during Stages~A1-A2, with the per-layer scaling described in Appendix~\ref{app:encoder-adapt}, and $\eta_{\text{enc}} = 0$ in Stage~A3.  A linear warm-up over $10{,}000$ steps is applied from $10^{-6}$ to each target value.

Gradient norms are clipped at $1.0$. Without clipping, sharp depth discontinuities in the 64-beam KITTI scans drive the gradient norm from $\approx\!2$ to $\approx\!42$ in a single step; no similar instability was observed on nuScenes.

Dropout rates: $p_{\text{mlp}}\!=\!0.1$, $p_{\text{attn}}\!=\!0.05$, $p_{\text{proj}}\!=\!0.1$.

EMA shadow weights are maintained at decay $0.9999$ for the DiT decoder and encoder, and at decay $0.999$ for the mask predictor. The faster mask EMA allows the shadow weights to track the rapidly evolving dynamics when the mask branch is introduced in Stage~F2 or Stage~A2.

Training uses 8$\times$A100-80\,GB GPUs, DDP, and BF16 mixed precision. Each GPU processes one 25-frame clip, but the model treats the frames as independent single-frame samples: no temporal attention, recurrent state, or temporal loss is used. Maximum training duration is 100 epochs.

\paragraph{Evaluation Protocol.} All point clouds are reconstructed in 3D Euclidean space at each dataset's native metric scale (metres). The F-score threshold of 0.5~m is therefore identical in absolute terms for both KITTI and nuScenes. Chamfer Distance and F-score are computed using the \texttt{ChamferDistancePytorch} library~\cite{chamfer2017}.

Results report mean $\pm$ standard deviation over \textbf{5 inference runs} that differ only in the stochastic seed used for the diffusion sampling and Bernoulli support sampling.  No retraining is performed between runs; each run initialises from the same final-stage checkpoint, namely Stage~F2 for frozen-encoder variants and Stage~A3 for encoder-adapted variants.

\paragraph{Fréchet metric protocol.} All Fréchet-style metrics (FRID, FSVD, FPVD) use every unique frame in the test split, with no subsampling: nuScenes provides $33{,}021$ real and $33{,}021$ generated frames after deduplicating overlapping 25-frame clips (the same frame count used in the step ablation, Fig.~\ref{fig:step-ablation}), and the KITTI-family provides $5{,}100$ real and $5{,}100$ generated frames. Each generated sample is the reconstruction paired with one real test frame, and means and covariances are computed over the complete feature sets. Feature extractors follow LiDM/LiDARGen: MinkowskiNet for FSVD, SPVCNN for FPVD, and RangeNet++ for FRID; the FRID checkpoint exists only for 64-beam range images, hence FRID is reported only for the KITTI-family (LiDM) setting.

\paragraph{Cross-seed stability.} Across the five reconstruction settings of Tables~\ref{tab:results_perceptual}--\ref{tab:results_geometry}, seed-to-seed standard deviation on edge F-score is at most $0.0002$ absolute (under $0.1\%$ relative) over the five runs that differ only in the diffusion and Bernoulli support-sampling seeds: sampling noise contributes almost nothing when the conditioning latent is fixed. In the plug-and-play world-model setting (Table~\ref{tab:results_world_model}), where latents are themselves generated rather than encoded from data, the spread rises to $0.0014$ for CRISP, but the deterministic baseline decoder shows a comparable $0.0017$ in the same setting: the increase comes from the sampled latents, not from CRISP's own sampling.

\section{Ablations}
The design ablations below use the SVD~$\times$3 setting on nuScenes, the most demanding frozen configuration due to the largest modality gap between a generic video VAE and sparse LiDAR geometry. This setting makes decoder failures most visible because the modality gap is largest; gains on LiDAR-native backbones may be smaller, and the LiDM replication at the end of this appendix (Table~\ref{tab:lidm-replication}) confirms that the same design choices transfer to the LiDAR-native LiDM backbone.

\begin{figure*}[h]
    \centering
    \includegraphics[width=0.99\linewidth]{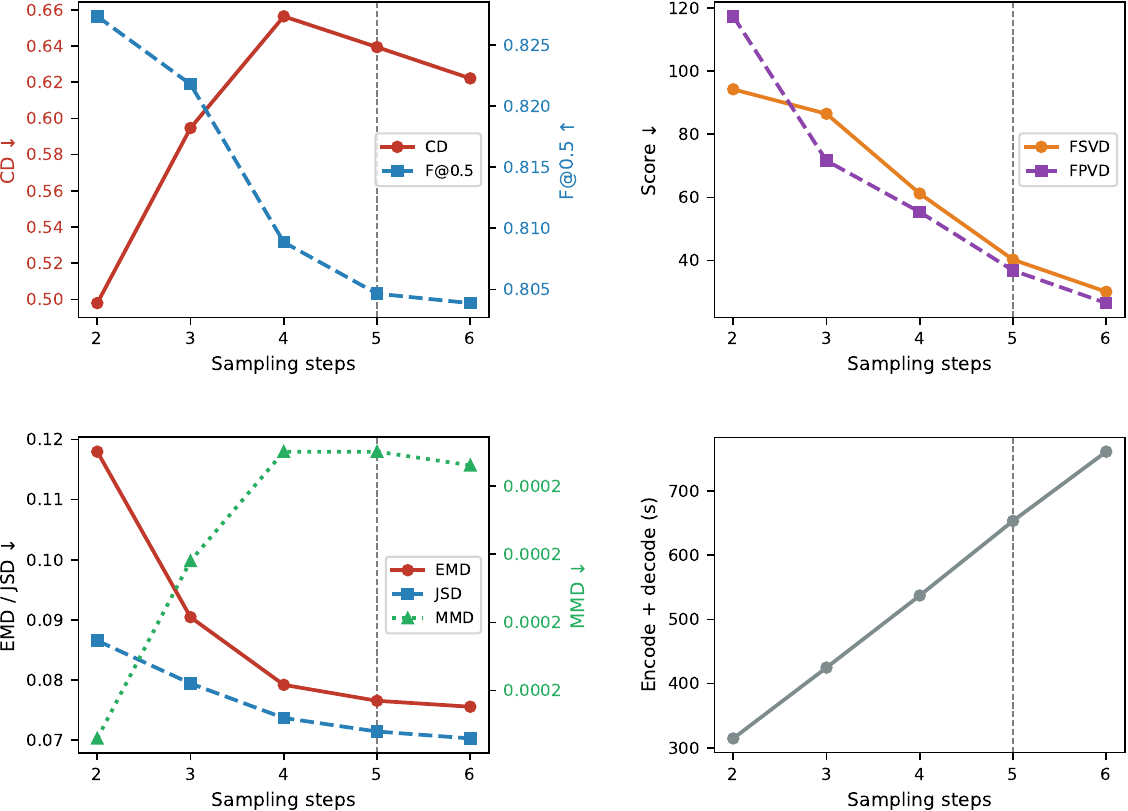}
    \caption{Step ablation on nuScenes with the SVD~$\times$3 backbone ($n\!=\!33{,}021$ LiDAR frames). Each panel plots a different metric group as a function of Euler sampling steps; the dashed vertical line marks the adopted 5-step setting. CD and perceptual metrics exhibit opposing trends, motivating a deliberate operating-point choice rather than optimising either metric alone.}
    \label{fig:step-ablation}
\end{figure*}

\paragraph{Diffusion Sampling Steps.}
\label{sec:ablation_appendix}
We ablate the number of Euler sampling steps over $n\!=\!33{,}021$ test-split LiDAR frames; results are shown in Fig.~\ref{fig:step-ablation}. The ablation reveals a Pareto trade-off: CD and perceptual quality pull in opposite directions across the step count, and the optimal operating point must balance both.

\emph{Geometric quality} (top-left): CD is lowest at 2 steps (CD$\!=\!0.499$) and degrades toward 4 steps (CD$\!=\!0.656$) before recovering at 5 (CD$\!=\!0.639$). The 2-step minimum reflects a known property of few-step diffusion: fewer denoising steps leave the output closer to the diffusion prior mean, which minimizes per-sample $\ell_2$-aligned metrics such as CD at the cost of perceptual sharpness. The non-monotone behaviour between 4 and 5 steps coincides with the sharpest perceptual improvement, suggesting the model reorganises its output distribution at this transition rather than simply refining it. F-score@0.5m decreases gradually with more steps (from $0.825$ at 2 steps to $0.804$ at 6), consistent with a small support miscalibration introduced as the denoiser sharpens boundary geometry.

\emph{Perceptual quality} (top-right): FSVD and FPVD decrease monotonically, from $94.2/117.3$ at 2 steps to $30.1/26.5$ at 6. The largest single drop occurs between 4 and 5 steps ($61.2\!\to\!40.3$ FSVD, $55.4\!\to\!36.9$ FPVD), confirming that 2-step outputs, despite their favourable CD, are perceptually poor and not representative of the real LiDAR distribution.

\emph{Statistical divergences} (bottom-left): MMD increases and plateaus after 4 steps, indicating a mild trade-off: additional denoising improves JSD/EMD and perceptual realism, but slightly worsens this nearest-set distributional metric.

\emph{Inference time} (bottom-right): encode+decode time grows approximately linearly, from 315\,s at 2 steps to 761\,s at 6 steps on 8$\times$H200 (without \texttt{torch.compile}; compiled throughput is in Table~\ref{tab:latency}).

We choose \emph{5 Euler steps} as the operating point: it sits past the perceptual phase transition (clearing most of the FSVD/FPVD gap), avoids the geometric regression at 4 steps, and adds only 30\,s over 4 steps. The residual 6-step improvement is marginal and not worth the additional latency.

\paragraph{DiT decoder design.}
Table~\ref{tab:ablation_dit} reports two independent comparisons on SVD~$\times$3.

\emph{Early fusion (upper group, GT mask, 35 epochs).} Removing early latent fusion degrades all metrics substantially: whole-cloud CD rises from $0.985$ to $1.134$ (+15\%), FSVD from $98.6$ to $119.6$ (+21\%), and edge CD from $7.244$ to $8.030$. Because the decoder starts from pure Gaussian noise with no auxiliary clean observation, the global scene layout must be established in the earliest transformer blocks; without early injection of latent tokens, the network can only infer layout from mid-level context, arriving too late to avoid locally incoherent geometry.

\emph{Explicit support prediction (lower group, converged).} Adding the mask branch reduces JSD by 35\% ($0.109\!\to\!0.071$), reduces EMD by 31\%, and improves edge F-score by 23\% ($0.193\!\to\!0.238$). The intuition is that depth values and binary valid returns are statistically incompatible signals: a single diffusion process that must simultaneously model continuous depth and binary support learns a compromised representation for both. The mask branch separates these two prediction problems, allowing the DiT to focus entirely on depth geometry. FPVD is the one metric where the no-mask variant edges out CRISP ($35.993$ vs $36.782$); FPVD is a volume-occupancy feature metric less sensitive to boundary precision than edge F-score or JSD, and the marginal gap is within the noise range observed across runs.

\begin{table*}[h]
\centering
\small
\setlength{\tabcolsep}{4pt}
\caption{DiT decoder ablation (SVD~$\times$3, nuScenes). \emph{Upper}: GT support mask applied, 35 epochs (3-run mean); early-fusion comparison. \emph{Lower}: converged end-to-end; mask-formulation comparison. Best within each group is bold.}
\label{tab:ablation_dit}
\vspace{0.5em}
\begin{tabular}{lrrrrrr}
\toprule
\multirow{2}{*}{Variant}
& \multicolumn{3}{c}{Statistical}
& \multicolumn{3}{c}{Perceptual} \\
\cmidrule(lr){2-4}\cmidrule(lr){5-7}
& JSD$\downarrow$ & EMD$\downarrow$ & MMD$\downarrow{\times}10^{-4}$
& FSVD$\downarrow$ & FPVD$\downarrow$ & FRID$\downarrow$ \\
\midrule
\multicolumn{7}{l}{\textbf{GT mask, 35 epochs}} \\
w/o early fusion & 0.120 & 0.241 & 3.101 & 119.557 & 100.839 & - \\
w/ early fusion     & \textbf{0.104} & \textbf{0.196} & \textbf{2.739} & \textbf{98.649} & \textbf{81.651}  & -\\
\multicolumn{7}{l}{\textbf{Full system, converged}} \\
Loss everywhere (no mask branch) & 0.109 & 0.110 & 2.371 & 42.556 & \textbf{35.993} & - \\
CRISP (ours)  & \textbf{0.071} & \textbf{0.076} & \textbf{2.251} & \textbf{40.302} & 36.782  & - \\
\bottomrule
\end{tabular} \par
\vspace{0.3em}
\begin{tabular}{lrrrrrr}
\multirow{2}{*}{Decoder}
& \multicolumn{2}{c}{Whole Geometry}
& \multicolumn{2}{c}{Edge Geometry}
& \multicolumn{2}{c}{Smooth Geometry} \\
\cmidrule(lr){2-3}\cmidrule(lr){4-5}\cmidrule(lr){6-7}
& CD$\downarrow$ & F$\uparrow$
& CD$\downarrow$ & F$\uparrow$
& CD$\downarrow$ & F$\uparrow$ \\
\midrule
\multicolumn{7}{l}{\textbf{GT mask, 35 epochs}} \\
w/o early fusion & 1.134 & 0.753 & 8.030 & 0.192 & 0.956 & 0.776 \\
w/ early fusion     & \textbf{0.985} & \textbf{0.757} & \textbf{7.244} & \textbf{0.199} & \textbf{0.825} & \textbf{0.780} \\
\multicolumn{7}{l}{\textbf{Full system, converged}} \\
Loss everywhere (no mask branch) & 0.660 & 0.797 & 8.083 & 0.193 & \textbf{0.537} & 0.819  \\
CRISP (ours)     & \textbf{0.639} & \textbf{0.804} & \textbf{7.364} &\textbf{ 0.238} & \textbf{0.537} & \textbf{0.821} \\
\bottomrule
\end{tabular}
\end{table*}

\paragraph{Mask predictor design.}
Table~\ref{tab:ablation_mask} ablates the mask predictor along two axes. The upper group holds the input fixed (depth\,+\,latent) and adds one loss term at a time, each trained for 10 epochs; the lower group holds the loss set fixed (full) and varies the input, both trained to convergence. The two groups use different training durations and are not directly comparable in absolute values.

\emph{Loss composition (upper group).} The largest single IoU gain is from adding Dice loss to Focal alone ($0.758\!\to\!0.827$, $+0.069$). This is expected: Focal BCE addresses per-pixel class imbalance but optimises a surrogate that does not directly measure overlap; Dice optimises overlap directly and is inherently recall-sensitive, recovering the many thin and sparse valid-return structures that Focal under-weights. Edge and Laplacian terms each add a further $+0.011$ and $+0.003$ respectively, capturing first-order boundary alignment and isolated ray-drop events that first-order gradients cannot detect. Hard-pixel mining adds $+0.004$ by up-weighting spatially fragile regions missed by the spatial-average losses. The full loss configuration with deep supervision (Stage 6) achieves IoU $0.844$, marginally below Stage 5's $0.845$; this gap is within single-run noise, but the full stack is retained because deep supervision ensures gradient signal reaches shallow encoder layers, improving robustness on thin-structure and high-azimuth-rate scenes beyond what the final IoU reflects.

\emph{Input conditioning (lower group).} Removing the latent branch and conditioning on depth alone drops converged IoU from $0.788$ to $0.630$ ($-20\%$) and F1 from $0.881$ to $0.773$. The depth hint provides high-frequency geometric cues, but cannot distinguish a true geometric void (sky, open road) from a sensor shadow or a sparse-return region caused by surface properties. The encoder latent carries semantic layout that resolves this ambiguity, enabling the predictor to correctly assign near-zero return probability to open voids and near-one probability to solid surfaces regardless of local depth texture.
\begin{table}[h]
\centering
\small
\setlength{\tabcolsep}{4pt}
\caption{Mask predictor ablation (SVD~$\times$3, nuScenes). \emph{Upper}: cumulative loss composition, depth\,+\,latent input, 10~ep.\ each. \emph{Lower}: input conditioning, full loss set, converged. $\dagger$~Stage~6 is the deployed configuration; the marginal gap vs.\ Stage~5 is within single-run noise but deep supervision is retained for robustness on thin-structure regions.}
\label{tab:ablation_mask}
\vspace{0.5em}
\begin{tabular}{lrrrr}
\toprule
Config & IoU$\uparrow$ & Prec.$\uparrow$ & Rec.$\uparrow$ & F1$\uparrow$ \\
\midrule
\multicolumn{5}{l}{\textbf{Loss composition (depth + latent, 10 ep.)}} \\
Focal only
  & 0.758 & 0.860 & 0.865 & 0.862 \\
\quad$+$ Dice
  & 0.827 & 0.864 & 0.951 & 0.906 \\
\quad$+$ Edge loss
  & 0.838 & 0.871 & 0.957 & 0.912 \\
\quad$+$ Laplacian
  & 0.841 & 0.869 & \textbf{0.963} & 0.914 \\
\quad$+$ Pixel mining
  & \textbf{0.845} & 0.874 & 0.962 & 0.916 \\
\quad$+$ Deep sup.\ (full)$^\dagger$
  & 0.844 & \textbf{0.874} & 0.961 & \textbf{0.917} \\
\midrule
\multicolumn{5}{l}{\textbf{Input conditioning (full losses, converged)}} \\
Depth only
  & 0.630 & 0.717 & 0.839 & 0.773 \\
Depth $+$ latent (ours)
  & \textbf{0.788} & \textbf{0.862} & \textbf{0.901} & \textbf{0.881} \\
\bottomrule
\end{tabular}
\end{table}

\paragraph{Cross-backbone replication (LiDM).} \label{sec:lidm_replication} The ablations above all use the frozen SVD~$\times$3 backbone. Table~\ref{tab:lidm-replication} reruns the two central design choices, latent-conditioning mechanism and Euler step count, on the frozen LiDAR-native LiDM backbone instead (24 blocks, 5 steps, 20 epochs unless varied; matched seed, data, schedule). The same choices win on LiDM: early+mid concatenation beats cross-attention and register tokens (on SVD~$\times$3 we instead ablated early fusion, Table~\ref{tab:ablation_dit}), and 5 Euler steps gives the lowest val L1, matching the operating point chosen on SVD~$\times$3 (Fig.~\ref{fig:step-ablation}). The mask-loss study behaves differently: LiDM support IoU starts at $0.9888$, and no single term moves it by more than $0.0011$ across a six-way leave-one-out (Table~\ref{tab:mask-loo}), so there is little left for the objective to recover once the LiDAR-native backbone is already near ceiling. The same pattern appears on the depth side, where removing importance weighting costs $25.2\%$ val L1 on nuScenes against only $5.9\%$ on LiDM/KITTI-family (Table~\ref{tab:depth-loo}). One confound remains: the LiDAR-native runs are also the 64-beam runs, so latent match and return density are entangled.

\begin{table}[h]
\centering\small
\caption{Design-choice replication on the frozen LiDAR-native LiDM backbone (KITTI-family). Validation L1 ($\downarrow$); cf.\ the SVD~$\times$3/nuScenes ablations in Table~\ref{tab:ablation_dit} and Fig.~\ref{fig:step-ablation}.}
\label{tab:lidm-replication}
\begin{tabular}{llc}
\toprule
Design axis & Variant & Val L1 $\downarrow$ \\
\midrule
\multirow{3}{*}{Latent conditioning} & Cross-attention & 0.1872 \\
 & Register tokens & 0.1156 \\
 & Early+mid concat (ours) & 0.0446 \\
\midrule
\multirow{4}{*}{Euler steps} & 1 & 0.0643 \\
 & 3 & 0.0527 \\
 & 5 (ours) & 0.0446 \\
 & 10 & 0.0789 \\
\bottomrule
\end{tabular}
\end{table}

\paragraph{Inference latency.}
Table~\ref{tab:latency} reports encoder and decoder throughput on 8$\times$\,H200 GPUs with \texttt{torch.compile}, batch size~25, and 5 diffusion sampling steps. The most notable result is Wan~$\times$3: CRISP's decoder (3.84\,ms/frame) is 2.5$\times$ faster than the inherited Wan2.1 VAE decoder (9.71\,ms/frame), because the original video VAE decoder is a full temporal video-decoding stack designed for high-resolution RGB video: a substantially heavier operation than CRISP's single-channel LiDAR-targeted DiT. The end-to-end CRISP pipeline on Wan~$\times$3 therefore runs at 86\,fps, faster than the baseline at 57\,fps, demonstrating that decoder replacement can simultaneously improve geometric quality and reduce latency when the inherited decoder is overbuilt for the target modality. For LiDM and SVD, CRISP's 5-step diffusion decoder adds 23.06\,ms and 3.92\,ms per frame respectively; both remain real-time capable (41 and 215\,fps) and the overhead is reducible via distillation~\cite{salimans2022progressive}.

\begin{table}[h]
\centering
\small
\setlength{\tabcolsep}{4pt}
\caption{End-to-end inference throughput (ms/frame and fps). Encoder cost is shared between baseline and CRISP. Measured on 8$\times$\,H200, \texttt{torch.compile}, batch~25, 5 Euler steps. $^\dagger$CRISP decoder is faster than the inherited Wan2.1 VAE decoder.}
\label{tab:latency}
\vspace{0.5em}
\begin{tabular}{lrrrrr}
\toprule
\multirow{2}{*}{Backbone}
& \multirow{2}{*}{\makecell{Encoder\\(ms/f)}}
& \multicolumn{2}{c}{Decoder (ms/f)}
& \multicolumn{2}{c}{End-to-end}\\
\cmidrule(lr){3-4}\cmidrule(lr){5-6}
&& Base. & CRISP & Base.\ fps & CRISP fps \\
\midrule
LiDM   & 1.31 & 1.39  & 23.06          & 369 & 41  \\
SVD~$\times$3 & 0.74 & 1.49  & \phantom{0}3.92 & 448 & 215  \\
Wan~$\times$3 & 7.78 & 9.71  & \phantom{0}3.84$^\dagger$ & \phantom{0}57 & \textbf{86} \\
\bottomrule
\end{tabular}
\end{table}

\paragraph{Decoder capacity: parameters and FLOPs.}
Table~\ref{tab:capacity} reports the parameter, FLOP, and latency comparison underlying the capacity controls discussed in the main text (paragraph ``Decoder capacity''). CRISP is essentially the same $\approx$516M network in every row (DiT decoder plus latent adapter $\approx$502M parameters, support-mask U-Net 13.8M; only the adapter's input width changes per backbone), so the ratios differ only through the size of the inherited decoder being replaced. Params and FLOPs are measured at batch~1, 5 Euler steps, and each setting's native range-map resolution ($32{\times}1024$ for SVD/Wan on nuScenes, $64{\times}1024$ for LiDM on KITTI-family), so the identical network performs $\approx2\times$ the FLOPs in the LiDM row. On LiDM, the same fixed-capacity step sweep as Fig.~\ref{fig:step-ablation} gives val L1 $0.0446$ (5 steps) rising to $0.0643$ (1 step) if iterative refinement is removed, i.e.\ $44\%$ higher error, with no change to parameters, conditioning, or data. One confound: LiDM's baseline decoder is already closest to its reconstruction ceiling, so part of its comparatively small gain (FSVD $-17.4\%$, FPVD $5.4\%$ worse) despite receiving the most extra capacity ($60\times$ params, $22.8\times$ FLOPs) is headroom rather than a capacity effect; SVD~$\times$3 and Wan~$\times$3 receive far less extra capacity ($7$--$8\times$ params, $3.4$--$3.7\times$ FLOPs) and improve the most (FSVD $-74.3\%$ and $-67.8\%$).

\section{Asset Licenses}
\label{app:licenses}

\begin{table}[h]
\centering
\small
\caption{Licenses for datasets, pretrained models, and third-party code used in this work.}
\begin{tabular}{lll}
\toprule
Asset & Reference & License \\
\midrule
KITTI-360        & \cite{liao2022kitti}        & CC BY-NC-SA 3.0 \\
SemanticKITTI    & \cite{behley2019semantickitti} & CC BY-NC-SA 4.0 \\
nuScenes         & \cite{caesar2020nuscenes}   & CC BY-NC-SA 4.0 \\
LiDM             & \cite{ran2024towards}       & MIT \\
SVD VAE          & \cite{blattmann2023stable}  & Stability AI Community License \\
Wan2.1 VAE       & \cite{wan2025wan}           & Apache 2.0 \\
ChamferDistancePytorch & \cite{chamfer2017}    & MIT \\
\bottomrule
\end{tabular}
\end{table}

\section{Extended Qualitative Results}
\label{sec:qual_appendix}

\begin{figure*}[t]
    \centering
    \includegraphics[width=0.99\linewidth]{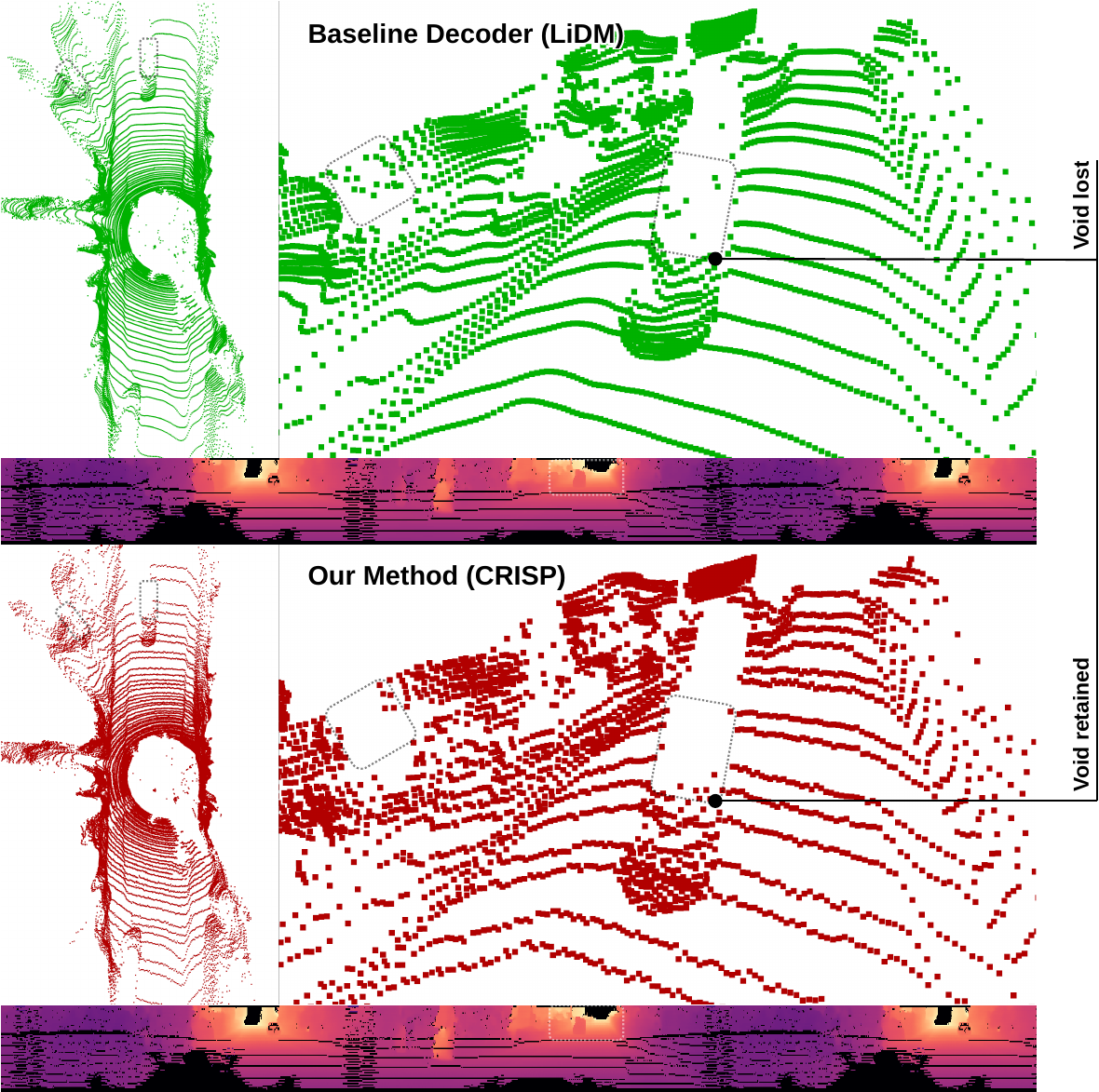}
    \caption{Extended qualitative comparison, LiDM world model (plug-and-play decoding, KITTI family). No ground truth is shown because these are generative samples from the frozen LiDM latent generator; both rows decode the same sampled latent. The baseline decoder introduces grid-aligned blocky artifacts and smears foreground boundaries. CRISP, swapped in without retraining the generator, produces cleaner scan geometry and sharper depth contours from the same latent input.}
    \label{fig:qual_extra_lidm_wm}
\end{figure*}

Except for the world-model samples in Fig.~\ref{fig:qual_extra_lidm_wm}, each figure shows the full-frame point cloud and the corresponding range map (horizontal strip); inset boxes indicate the cropped region shown on the right. Colors follow Fig.~\ref{fig:intro_figure}: \textcolor{blue}{blue} is ground truth, \textcolor{green!50!black}{green} is the inherited baseline decoder, and \textcolor{red!70!black}{red} is CRISP.

\begin{figure*}[t]
    \centering
    \includegraphics[width=0.99\linewidth]{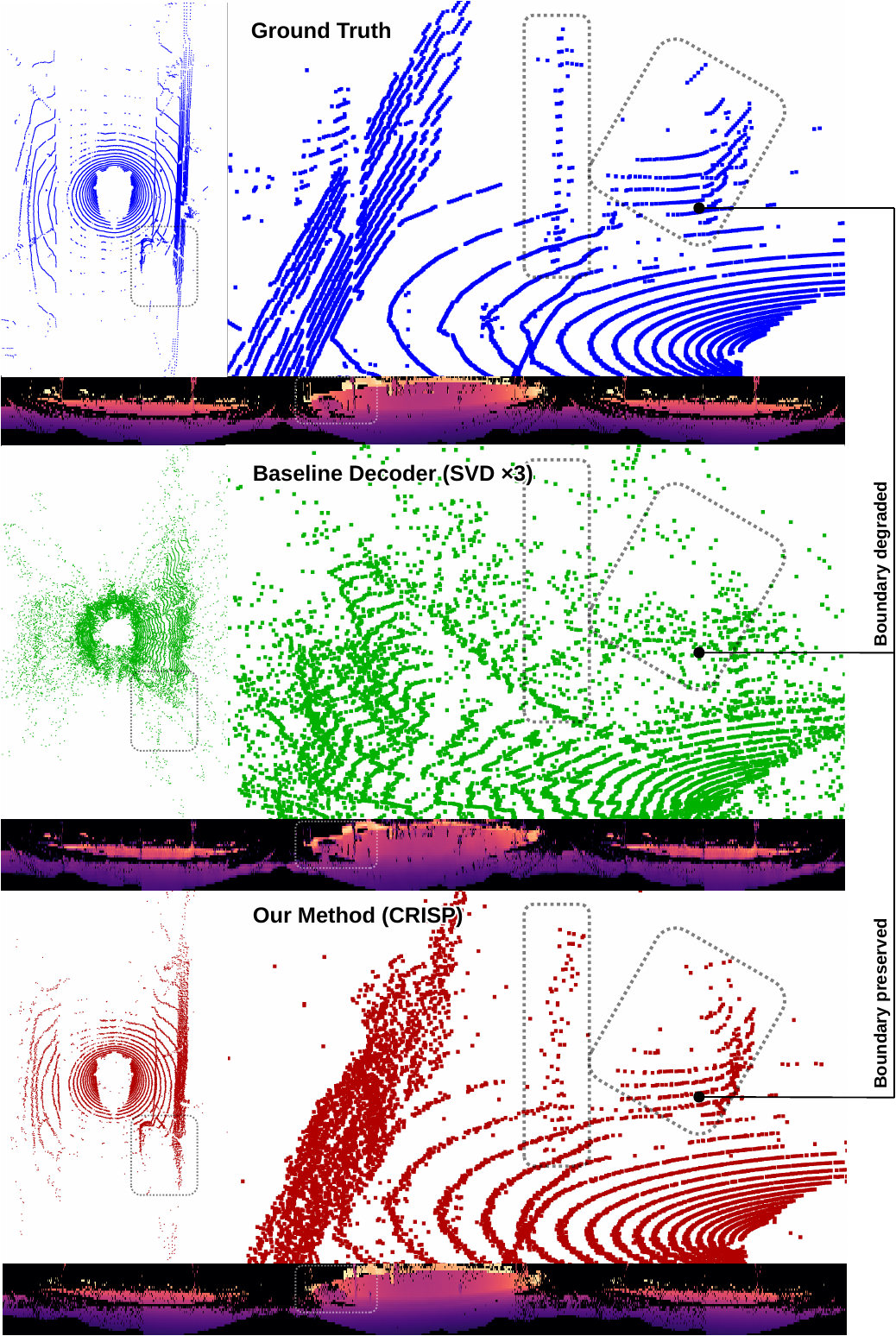}
    \caption{Extended qualitative comparison, SVD~$\times$3 (frozen encoder, nuScenes). The baseline decoder introduces heavy scan-line scatter at depth discontinuities: beam-level structure of a vehicle and pole, visible in ground truth, dissolves into noise clouds in the baseline reconstruction. CRISP restores continuous scan-line geometry and suppresses the boundary-bridging artifacts highlighted by the inset box.}
    \label{fig:qual_extra_svd3x}
\end{figure*}

\begin{figure*}[t]
    \centering
    \includegraphics[width=0.99\linewidth]{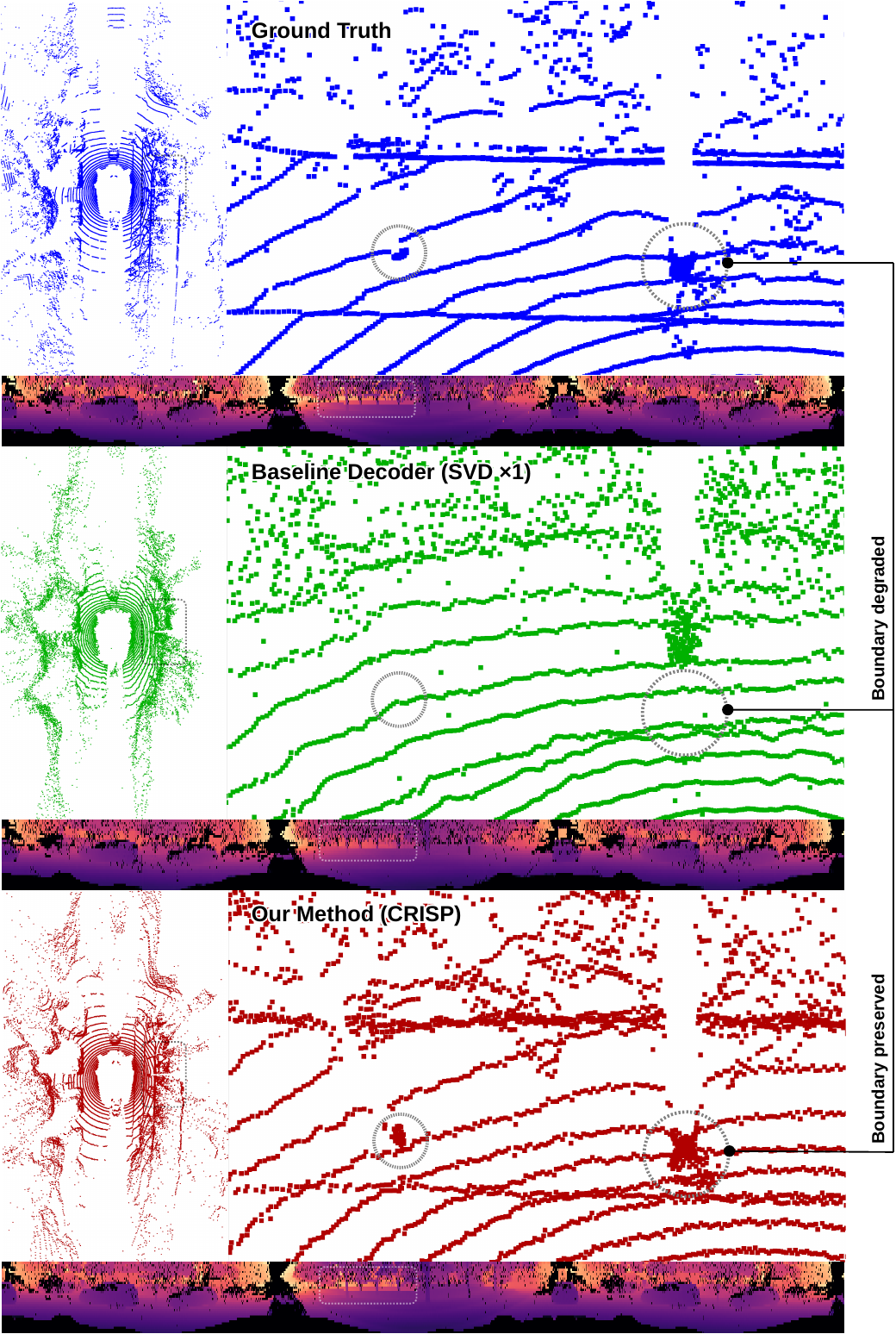}
    \caption{Extended qualitative comparison, SVD~$\times$1 (encoder-adapted, nuScenes). The baseline decoder misplaces a mid-range object cluster (circled tree trunks): points smear outward rather than forming a compact boundary. CRISP localises the cluster accurately and sharpens the depth contour at the foreground-background transition indicated by the inset box.}
    \label{fig:qual_extra_svd1x}
\end{figure*}

\begin{figure*}[t]
    \centering
    \includegraphics[width=0.99\linewidth]{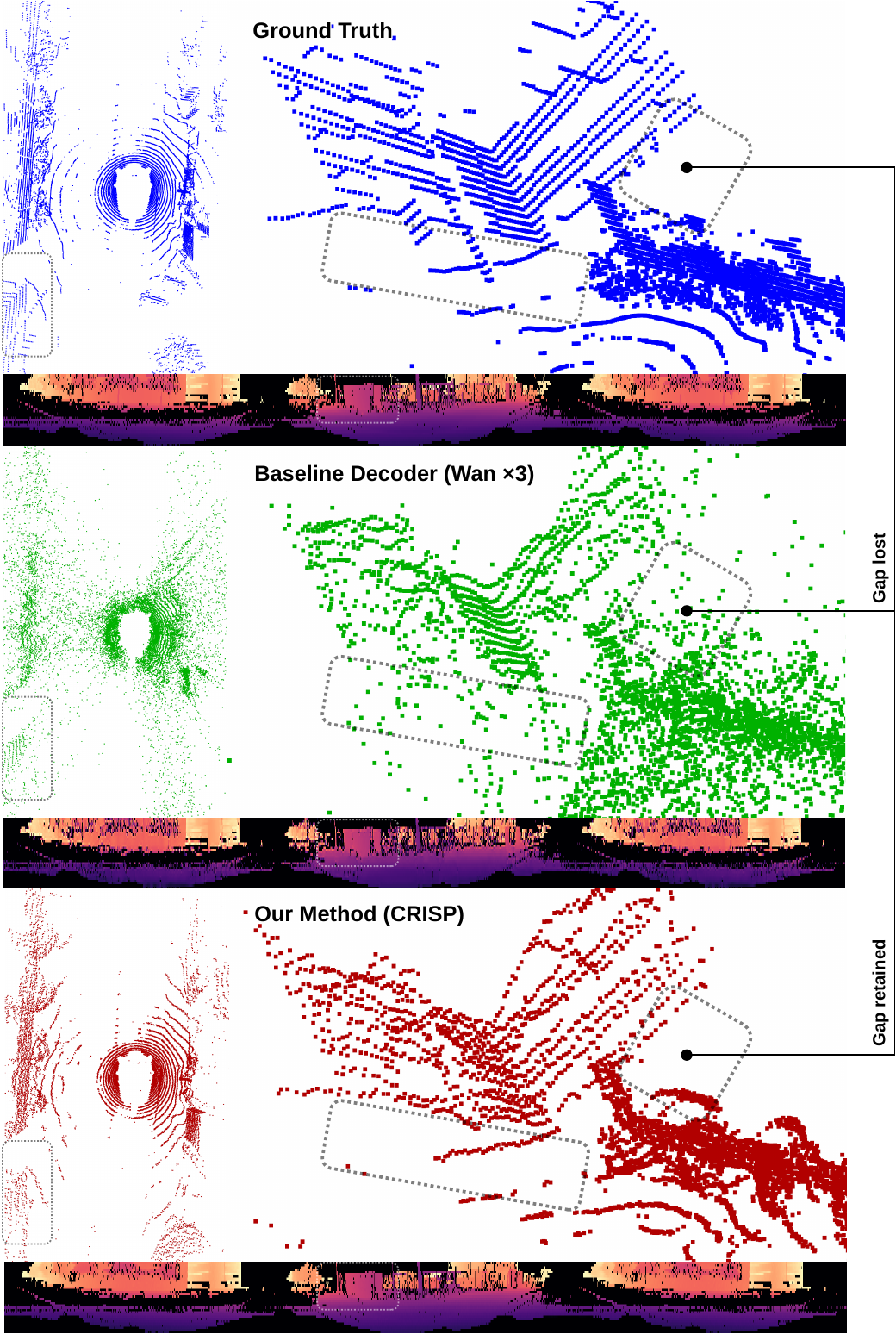}
    \caption{Extended qualitative comparison, Wan~$\times$3 (frozen encoder, nuScenes). The highlighted region contains a foreground-background scene depth gap that the baseline decoder bridges: points fill in between the two surfaces, erasing the geometric discontinuity. CRISP retains the gap as a sharp void, consistent with the ground-truth scan.}
    \label{fig:qual_extra_wan3x}
\end{figure*}

\begin{figure*}[t]
    \centering
    \includegraphics[width=0.99\linewidth]{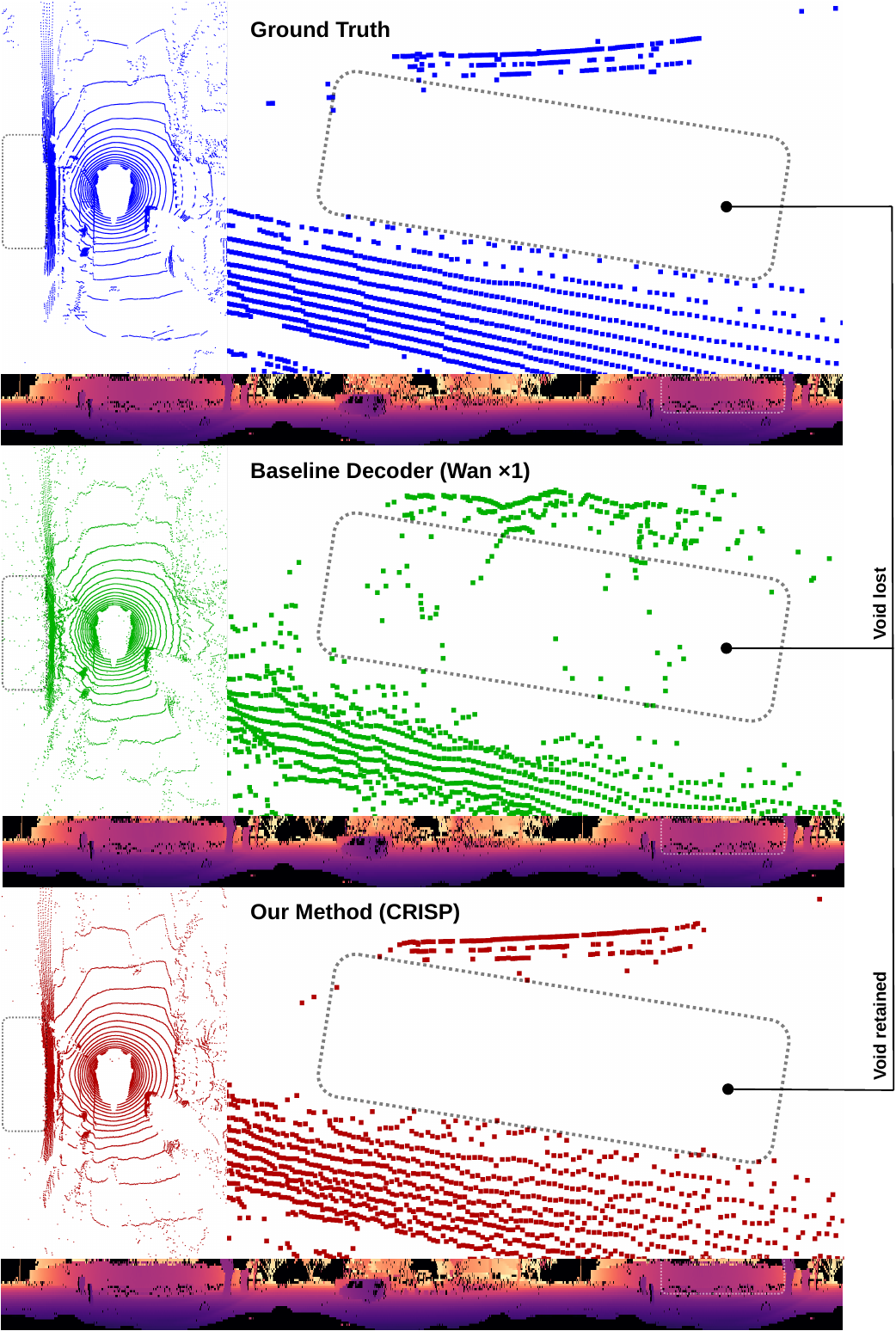}
    \caption{Extended qualitative comparison, Wan~$\times$1 (encoder-adapted, nuScenes). The baseline decoder incorrectly fills the open void region visible in ground truth (inset box): spurious returns appear where no surface exists. CRISP preserves the void, producing a sparse valid-return map that matches the support structure of the real scan.}
    \label{fig:qual_extra_wan1x}
\end{figure*}

\begin{figure*}[t]
    \centering
    \includegraphics[width=0.99\linewidth]{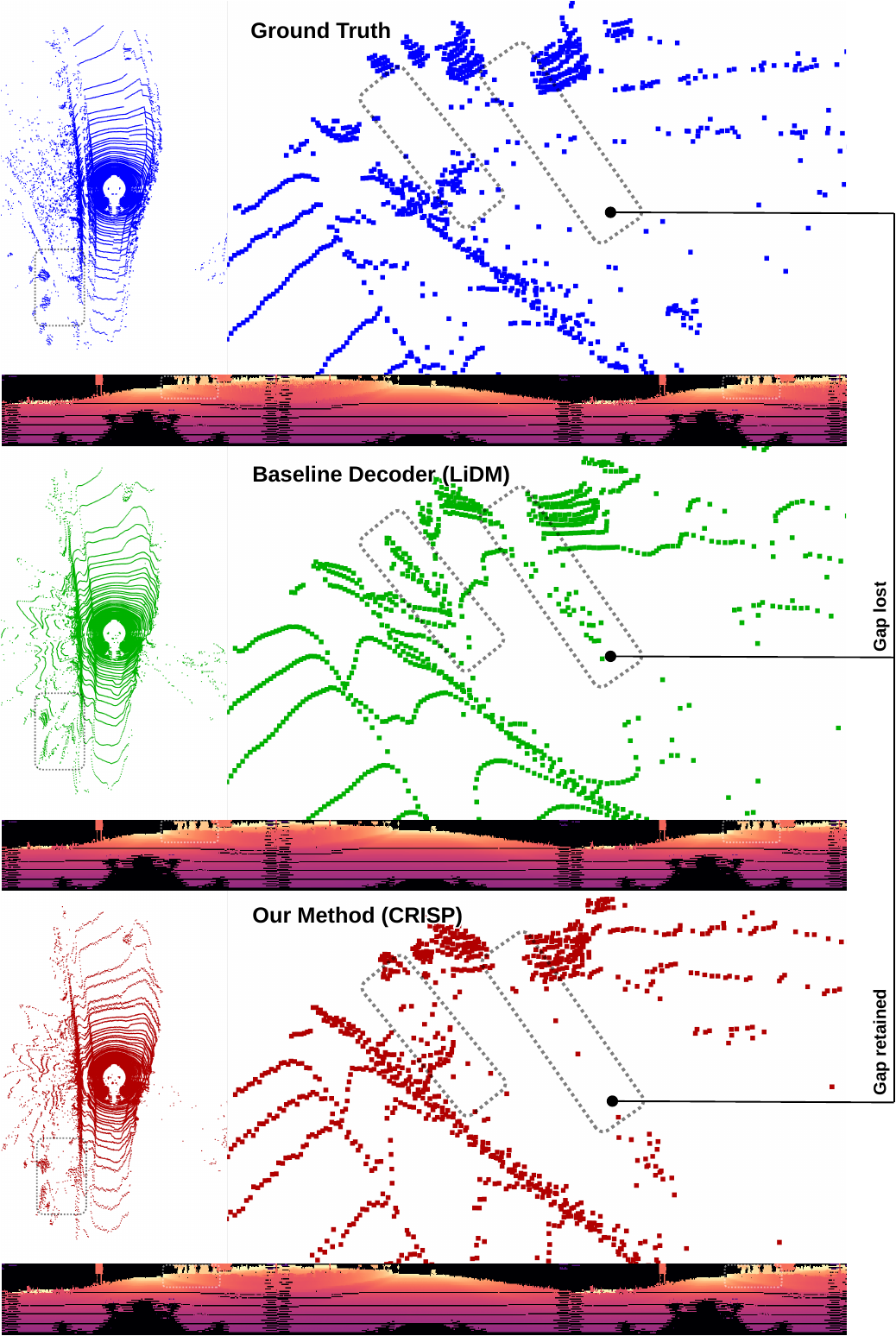}
    \caption{Extended qualitative comparison, LiDM (frozen encoder, KITTI family). The 64-beam KITTI geometry makes foreground-background depth gaps particularly sharp. The baseline decoder bridges the gap in the highlighted region, collapsing two geometrically distinct surfaces. CRISP reproduces the correct depth discontinuity, matching the gap geometry visible in the ground-truth scan.}
    \label{fig:qual_extra_lidm}
\end{figure*}

\clearpage

\clearpage
\newpage
\section*{NeurIPS Paper Checklist}

\begin{enumerate}

\item {\bf Claims}
    \item[] Question: Do the main claims made in the abstract and introduction accurately reflect the paper's contributions and scope?
    \item[] Answer: \answerYes{}
    \item[] Justification: The abstract and introduction claim a plug-in pixel-space LiDAR decoder that works across multiple latent backbones and improves reconstruction, especially near discontinuities. These claims match the method and results sections. The paper also discusses non-uniform behavior across metrics, including perceptual or distributional regressions in some adapted and plug-and-play settings, rather than claiming uniform superiority.

\item {\bf Limitations}
    \item[] Question: Does the paper discuss the limitations of the work performed by the authors?
    \item[] Answer: \answerYes{}
    \item[] Justification: A dedicated Limitations paragraph at the end of the Results section discusses: (1) operation on a single depth channel without remission or intensity; (2) reliance on range-map assumptions shared by prior latent LiDAR pipelines; (3) residual high-frequency noise on smooth surfaces; (4) possible thin-structure flicker due to independent frame decoding; and (5) inference overhead of 3.9--23\,ms/frame for SVD and LiDM, with distillation noted as a mitigation path.

\item {\bf Theory assumptions and proofs}
    \item[] Question: For each theoretical result, does the paper provide the full set of assumptions and a complete (and correct) proof?
    \item[] Answer: \answerNA{}
    \item[] Justification: The paper does not present formal theoretical results such as theorems, lemmas, or proofs. Its equations define the model and training objectives rather than results requiring mathematical proof.

    \item {\bf Experimental result reproducibility}
    \item[] Question: Does the paper fully disclose all the information needed to reproduce the main experimental results of the paper to the extent that it affects the main claims and/or conclusions of the paper (regardless of whether the code and data are provided or not)?
    \item[] Answer: \answerYes{}
    \item[] Justification: The method section specifies the model forward pass, adapter, diffusion decoder, support mask predictor, and main optimization objectives. The appendix provides architecture details, dataset preprocessing, splits, augmentations, optimizer settings, learning rates, training stages, EMA, dropout, gradient clipping, loss weights, and inference protocol. Main tables report mean and standard deviation over five stochastic inference runs from a fixed checkpoint. The boundary-focused evaluation defines edge pixels by thresholding valid-only local depth discontinuities, with the threshold value reported in the Experimental Setup section and implementation details provided in the appendix.

\item {\bf Open access to data and code}
    \item[] Question: Does the paper provide open access to the data and code, with sufficient instructions to faithfully reproduce the main experimental results, as described in supplemental material?
    \item[] Answer: \answerNo{}
    \item[] Justification: Code and checkpoints are not yet publicly released. All evaluation datasets are publicly available under their respective licenses, and the pretrained backbones are publicly available. The paper provides detailed training and evaluation specifications to support reproducibility, and code, training scripts, evaluation scripts, and CRISP checkpoints will be released on the project page (\url{https://andrea25512.github.io/CRISP/}).

\item {\bf Experimental setting/details}
    \item[] Question: Does the paper specify all the training and test details (e.g., data splits, hyperparameters, how they were chosen, type of optimizer) necessary to understand the results?
    \item[] Answer: \answerYes{}
    \item[] Justification: The Experimental Setup section specifies datasets, split sizes, range-map resolutions, latent backbones, comparison regimes (frozen vs.\ encoder-adapted vs.\ plug-and-play), and all reported metrics with their definitions. Appendices~C--D provides full training hyperparameters, augmentation protocol, diffusion schedule, and the backbone-specific adapter configurations used for each setting.

\item {\bf Experiment statistical significance}
    \item[] Question: Does the paper report error bars suitably and correctly defined or other appropriate information about the statistical significance of the experiments?
    \item[] Answer: \answerYes{}
    \item[] Justification: All main quantitative tables report mean and standard deviation over five stochastic inference runs that differ only in the random seed used for diffusion sampling and Bernoulli support sampling. No retraining is performed between runs; each run uses the same final-stage checkpoint, namely Stage~F2 for frozen-encoder variants and Stage~A3 for encoder-adapted variants. The reported variability therefore reflects sampling stochasticity rather than training-seed variability.

\item {\bf Experiments compute resources}
    \item[] Question: For each experiment, does the paper provide sufficient information on the computer resources (type of compute workers, memory, time of execution) needed to reproduce the experiments?
    \item[] Answer: \answerYes{}
    \item[] Justification: The appendix reports the GPU type and memory used for training (8$\times$A100-80GB), mixed-precision/DDP setup, batch structure, optimizer configuration, and approximate CRISP training cost: about 36~h per full run, consisting of roughly 24~h for the DiT-only stage and 12~h after introducing the support mask branch, corresponding to about 288 GPU-hours per run. It also reports inference throughput on 8$\times$H200 GPUs with \texttt{torch.compile}, batch size 25, and 5 Euler steps. Baseline VAE adaptation and ablation runtimes are excluded from the reported training-cost estimate.
    
\item {\bf Code of ethics}
    \item[] Question: Does the research conducted in the paper conform, in every respect, with the NeurIPS Code of Ethics \url{https://neurips.cc/public/EthicsGuidelines}?
    \item[] Answer: \answerYes{}
    \item[] Justification: The work is an offline methodological study on LiDAR reconstruction using publicly licensed automotive LiDAR datasets and pretrained models. It does not involve human subjects, deceptive experimentation, or private/sensitive data collection. To the authors' knowledge, the study conforms to the NeurIPS Code of Ethics.

\item {\bf Broader impacts}
    \item[] Question: Does the paper discuss both potential positive societal impacts and negative societal impacts of the work performed?
    \item[] Answer: \answerYes{}
    \item[] Justification: The intended positive impact is improved synthetic LiDAR fidelity for autonomous-driving simulation, which may support safer perception and planning evaluation and reduce reliance on costly real-world collection. Potential negative impacts include misuse of high-fidelity synthetic LiDAR to generate deceptive sensor traces, spoof perception systems, or create misleading safety-audit evidence. These risks are moderated by the domain-specific nature of CRISP, which is a decoder component rather than a standalone broad generative model, but they remain relevant for responsible release and downstream use.
    
\item {\bf Safeguards}
    \item[] Question: Does the paper describe safeguards that have been put in place for responsible release of data or models that have a high risk for misuse (e.g., pre-trained language models, image generators, or scraped datasets)?
    \item[] Answer: \answerNA{}
    \item[] Justification: CRISP is a domain-specific decoder component for automotive LiDAR range maps and is not a broad content-generation model, scraped dataset, or high-risk pretrained foundation model. The anticipated release consists of code and checkpoints trained on publicly licensed automotive LiDAR datasets, so dedicated release safeguards beyond standard licensing, documentation, and responsible-use notes are not applicable.

\item {\bf Licenses for existing assets}
    \item[] Question: Are the creators or original owners of assets (e.g., code, data, models), used in the paper, properly credited and are the license and terms of use explicitly mentioned and properly respected?
    \item[] Answer: \answerYes{}
    \item[] Justification: All datasets and pretrained models used in experiments are publicly available and cited. Full license information for each asset is provided in Appendix~\ref{app:licenses}.

\item {\bf New assets}
    \item[] Question: Are new assets introduced in the paper well documented and is the documentation provided alongside the assets?
    \item[] Answer: \answerNo{}
    \item[] Justification: The paper introduces new CRISP decoder checkpoints, latent adapter weights, and training/evaluation code, but these assets are not yet publicly released. The release on the project page will include documentation covering training data, intended use, limitations, licenses, dependencies, and reproduction instructions.

\item {\bf Crowdsourcing and research with human subjects}
    \item[] Question: For crowdsourcing experiments and research with human subjects, does the paper include the full text of instructions given to participants and screenshots, if applicable, as well as details about compensation (if any)? 
    \item[] Answer: \answerNA{}
        \item[] Justification: The paper does not involve crowdsourcing, surveys, manual human-subject studies, or compensated participant labor. Therefore instructions to participants, screenshots, and compensation details are not applicable.

\item {\bf Institutional review board (IRB) approvals or equivalent for research with human subjects}
    \item[] Question: Does the paper describe potential risks incurred by study participants, whether such risks were disclosed to the subjects, and whether Institutional Review Board (IRB) approvals (or an equivalent approval/review based on the requirements of your country or institution) were obtained?
    \item[] Answer: \answerNA{}
    \item[] Justification: The work does not involve human subjects or participant studies. Accordingly, IRB approval and participant-risk disclosures are not applicable.

\item {\bf Declaration of LLM usage}
    \item[] Question: Does the paper describe the usage of LLMs if it is an important, original, or non-standard component of the core methods in this research? Note that if the LLM is used only for writing, editing, or formatting purposes and does \emph{not} impact the core methodology, scientific rigor, or originality of the research, declaration is not required.
    \item[] Answer: \answerNA{}
    \item[] Justification: LLMs are not part of the core method, experiments, or scientific claims in this paper. Under the checklist guidance, writing or editing assistance alone does not require declaration here.

\end{enumerate}

\end{document}